\documentclass{article}
\PassOptionsToPackage{numbers,sort}{natbib}
\usepackage{journal}
\jmlrheading{}{2026}{}{}{}{O.~Montasser}
\makeatletter
\def\ps@jmlrtps{\let\@mkboth\@gobbletwo
  \def\@oddhead{}\def\@evenhead{}\def\@evenfoot{}%
  \def\@oddfoot{\small\copyright\,2026 O.~Montasser.\hfill To appear in NeurIPS 2026.}}
\makeatother
\ShortHeadings{Agnostic Online Learning with Reliable Abstention}{Montasser}
\usepackage{times}
\usepackage[utf8]{inputenc}
\usepackage[T1]{fontenc}
\usepackage{amsmath,amssymb,amsthm,mathtools}
\usepackage{algorithm}
\usepackage[algo2e,ruled,linesnumbered,vlined]{algorithm2e}
\usepackage[noend]{algpseudocode}
\usepackage{microtype}
\usepackage[dvipsnames]{xcolor}
\usepackage{hyperref}
\usepackage{prettyref}
\usepackage{nicefrac}
\usepackage{enumitem}
\hypersetup{colorlinks=true,citecolor=Maroon,linkcolor=Maroon,urlcolor=Maroon}
\setcitestyle{numbers,square}
\usepackage{comment}
\usepackage{dsfont}
\usepackage{thm-restate}
\usepackage{booktabs}

\newrefformat{alg}{\hyperref[#1]{Algorithm~\ref*{#1}}}
\newrefformat{thm}{\hyperref[#1]{Theorem~\ref*{#1}}}
\newrefformat{corr}{\hyperref[#1]{Corollary~\ref*{#1}}}
\newrefformat{lem}{\hyperref[#1]{Lemma~\ref*{#1}}}
\newrefformat{def}{\hyperref[#1]{Definition~\ref*{#1}}}
\newrefformat{eqn}{\hyperref[#1]{(\ref*{#1})}}
\newrefformat{rem}{\hyperref[#1]{Remark~\ref*{#1}}}
\newrefformat{sec}{\hyperref[#1]{Section~\ref*{#1}}}
\newrefformat{app}{\hyperref[#1]{Appendix~\ref*{#1}}}
\newrefformat{tab}{\hyperref[#1]{Table~\ref*{#1}}}

\newtheorem{theorem}{Theorem}
\newtheorem{lemma}{Lemma}
\newtheorem{corollary}{Corollary}

\newcommand{\restatedtheoremname}{}
\newtheorem*{restatedtheoreminner}{\protect\restatedtheoremname}
\newenvironment{restatedtheorem}[1]{\renewcommand{\restatedtheoremname}{#1}\begin{restatedtheoreminner}}{\end{restatedtheoreminner}}
\theoremstyle{definition}
\newtheorem{definition}{Definition}
\newtheorem{remark}{Remark}

\newcommand{\X}{\mathcal X}
\newcommand{\Hc}{\mathcal H}
\newcommand{\Bc}{\mathcal B}

\newcommand{\Gc}{\mathcal G}
\newcommand{\Fc}{\mathcal F}
\newcommand{\R}{\mathbb R}
\newcommand{\E}{\mathds{E}}
\newcommand{\Prob}{\mathbb P}
\newcommand{\1}{\mathds 1}
\newcommand{\Mis}{\mathrm{MissErr}}
\newcommand{\Abs}{\mathrm{AbsErr}}
\newcommand{\OPT}{\mathrm{OPT}}
\newcommand{\OPTc}{\mathrm{OPT}_{\mathrm{cen}}}

\newcommand{\iid}{\stackrel{\mathrm{iid}}{\sim}}

\newcommand{\Nbr}[3]{\mathcal N_{[~]}(#1,#2,#3)}

\newcommand{\eps}{\varepsilon}
\newcommand{\prn}[1]{\left(#1\right)}
\newcommand{\brk}[1]{\left[#1\right]}
\newcommand{\set}[1]{\left\{#1\right\}}
\newcommand{\ceil}[1]{\left\lceil#1\right\rceil}

\title{Agnostic Online Learning with Reliable Abstention}

\author{%
  \name{Omar Montasser} \email{omar.montasser@yale.edu}\\
  \addr Yale University\\
}

\begin{document}

\maketitle

\begin{abstract}
  We study online learning in the adversarial injection model \citep*{DBLP:conf/nips/GoelHMS23}, where each instance is either drawn i.i.d.~from an unknown distribution or is an arbitrary adversarial injection. The learner never knows which rounds are adversarial, but may abstain without penalty on adversarial injections. The learner's goal is to make few total misclassifications and few abstentions on i.i.d.~instances. This model lies between classical statistical learning, governed by the VC dimension, and adversarial online learning, governed by the Littlestone dimension. Unlike prior work, which has largely focused on the realizable setting, we consider an agnostic formulation that allows arbitrary labels. We present an online algorithm that guarantees sublinear excess misclassification error and sublinear abstention error, against oblivious and adaptive adversaries, for function classes that admit bounded bracketing number witnessed by brackets of bounded VC dimension. Our algorithm is distribution-independent and does {\em not} require prior knowledge of the distribution. As corollaries, we obtain new learning guarantees for halfspaces (linear classifiers) and richer classes such as feed-forward neural networks with threshold activations.
\end{abstract}

\section{Introduction}

A stark gap arises in learning theory when considering online (or sequential) classification between the fully-stochastic and fully-adversarial worlds. In the stochastic world, where the online data sequence is independent and identically distributed (i.i.d.) from some (unknown) distribution $D$, it is well-known that a function class $\Hc$ is learnable if and only if its VC dimension is finite \cite{VapnikChervonenkis71, DBLP:journals/jcss/LazaricM12}. In the adversarial world, where the online data sequence is adversarially and adaptively chosen, a function class $\Hc$ is learnable if and only if its Littlestone dimension is finite \cite{DBLP:journals/ml/Littlestone87, DBLP:conf/colt/Ben-DavidPS09}. The stark gap between these two worlds is exemplified by arguably one of the simplest function classes: thresholds on the unit interval, which has VC dimension 1 but {\em infinite} Littlestone dimension.

This gap continues to inspire and motivate research that explores intermediate relaxations between the two worlds, with the aim of obtaining strong learning guarantees (as in the stochastic world) but under more general conditions (as in the adversarial world). For example, this includes research on smoothed online learning \cite{DBLP:conf/nips/RakhlinST11, DBLP:journals/jacm/HaghtalabRS24, DBLP:conf/colt/BlockDGR22}, distributionally-constrained adversaries \citep*{DBLP:journals/corr/abs-2506-10293, DBLP:journals/corr/abs-2602-20585}, transductive online learning \citep*{BKM97, HMS23}, and introducing relaxed benchmarks to compete with \citep*{DBLP:conf/colt/MontasserSZ25}.

The focus of this work is on one particularly relevant model introduced by Goel et al.~\cite{DBLP:conf/nips/GoelHMS23} for beyond worst-case online classification. This model considers online classification of i.i.d.~sequences interspersed with arbitrary adversarial injections where the learner is allowed to {\em abstain}. Specifically, in each time step $t=1,\dots, T$, the adversary decides whether: (a) $x_t$ is drawn i.i.d.~from an (unknown) distribution $D$, or (b) $x_t$ is an adversarial injection chosen arbitrarily from $\X$. The adversary's decisions are never revealed to the learner. Importantly, the learner is allowed to {\em abstain} on adversarial injections, but is penalized for abstaining on i.i.d.~instances. The goal of the learner is to make as few misclassifications as possible, and as few abstentions as possible on the i.i.d.~subsequence,
\begin{equation}
  \label{eqn:obj}
  \Mis:=\sum_{t=1}^T \1\bigl[\hat y_t\in\{0,1\}\text{ and }\hat y_t\neq y_t\bigr] \text{ and } \Abs:=\sum_{t=1}^T \1\bigl[\hat y_t=\perp\text{ and }x_t\text{ is i.i.d.}\bigr].
\end{equation}

Prior work on this problem has mainly focused on the {\em realizable} setting where the ground-truth labels of the i.i.d.~sequence {\em and} the adversarial injections are consistent with some target function $h_\star \in \Hc$, i.e., $y_t = h_\star(x_t)$ \citep*{DBLP:conf/nips/GoelHMS23, YB26, EdelmanGoel26}. In this work, we consider the more challenging {\em agnostic} setting of this problem, and by that we mean that the labels of the i.i.d.~sequence and the adversarial injections are {\em arbitrarily} chosen by the adversary and need not be consistent with some $h_\star \in \Hc$. The central question that we study is:

\begin{center}
  \textit{What learning guarantees can we hope to achieve for {\em agnostic} online classification\\with abstention under adversarial injections?}
\end{center}
As is standard in agnostic online learning \citep*{DBLP:conf/colt/Ben-DavidPS09}, a sensible objective is to require the learner to achieve low misclassification error ($\Mis$) relative to the best classifier $h_\star \in \Hc$,
\begin{equation}
  \label{eqn:agnostic}
  \Mis - \OPT = \sum_{t=1}^T \1\bigl[\hat y_t\in\{0,1\}\text{ and }\hat y_t\neq y_t\bigr] - \min_{h\in \Hc} \sum_{t=1}^{T} \1[h(x_t) \neq y_t],
\end{equation}
while simultaneously requiring low abstention error ($\Abs$) as defined in \prettyref{eqn:obj}. In fact, we will also consider competing with a {\em stricter} benchmark, $\OPTc$, which is the best-in-class error on the (subset) rounds the learner did {\em not} abstain (see \prettyref{rem:cens}),
\begin{equation}
  \label{eqn:optc-intro}
  \OPTc := \min_{h \in \Hc} \sum_{t=1}^{T} \1[\hat y_t \neq \perp] \cdot\1[h(x_t)\neq y_t].
\end{equation}

Before presenting our contributions, we put in context some of the results in the realizable setting ($\OPT=0$). It is important to distinguish between two cases: (i) the learner has prior knowledge of distribution $D$ (distribution-dependent), and (ii) the learner has {\em no} prior knowledge of $D$ (distribution-independent). With prior knowledge of $D$, Goel et al.~\cite{DBLP:conf/nips/GoelHMS23} proposed a learning algorithm that achieves a total misclassification error of $O(d^2\log(1/\eps))$ and abstention error of $O(\eps T)$, where $d$ denotes the VC dimension of $\Hc$. More recently, in the more challenging case of {\em no} prior knowledge of $D$, Yu and Blanchard~\cite{YB26} proposed a learning algorithm that achieves misclassification error of $O(T^{3\alpha})$ and abstention error of $O(d^2\log^{\nicefrac53}(T)\cdot T^{1-\alpha})$ for any $\alpha \in [0,\nicefrac13]$ against {\em oblivious} adversaries. Furthermore, Yu and Blanchard~\cite{YB26} extended those guarantees to {\em adaptive} adversaries for function classes satisfying additional structure that they introduce (finite reduction dimension) which includes halfspaces. Another important result is that, without prior knowledge of $D$ (i.e., distribution-independent), polynomial tradeoffs between misclassification and abstention errors are {\em unavoidable} even for oblivious adversaries \cite{EdelmanGoel26, YB26}. Specifically, there are function classes with VC dimension 1 such that any algorithm must incur either $\Omega(\sqrt{T})$ misclassification error or $\Omega(\sqrt{T})$ abstention error \citep*{EdelmanGoel26, YB26}. Beyond the realizable setting, no learning guarantees were known prior to this work except perhaps for the special case of thresholds under random classification noise \cite{DBLP:journals/corr/abs-2504-13966}.
\subsection{Our Contributions}
In this paper, we study the problem of online classification of i.i.d.~sequences interspersed with arbitrary adversarial injections introduced by \cite{DBLP:conf/nips/GoelHMS23}, with a particular focus on obtaining {\em agnostic} guarantees making no assumptions on how the labels are generated.

\paragraph{Results.} Our formal results are presented in \prettyref{sec:mainresult} and summarized in \prettyref{tab:results}. We provide an overview below:
\begin{itemize}[leftmargin=1.5em]
  \item In \prettyref{thm:main-result}, we present an algorithm (\prettyref{alg:agnostic}) that achieves sublinear misclassification and abstention errors in the {\em agnostic} and realizable settings, against oblivious {\em and} adaptive adversaries, for function classes $\Hc$ that admit bounded bracketing number witnessed by brackets of bounded VC dimension. Our algorithm is distribution-independent and does {\em not} require prior knowledge of the underlying distribution.
  \item In \prettyref{thm:censored}, we extend quantitatively similar learning guarantees to a {\em censored} variant of the problem where the learner observes feedback labels only in the rounds where it does {\em not} abstain and is required to compete with the best-in-class on these rounds.
  \item In \prettyref{corr:halfspaces}, we instantiate our generic guarantees to establish new and improved results for the class of halfspaces in $\mathbb{R}^d$. In \prettyref{corr:comp-halfspaces}, we establish new learning guarantees for richer function classes such as intersections of halfspaces and feed-forward neural networks. Along the way, we establish a new generic structural result bounding the $\eps$-bracketing number of compositions of classes (\prettyref{lem:composition-bracketing}) which may be of independent interest.
\end{itemize}

\begin{table}[H]
  \centering
  \small
  \begin{tabular}{llll}
    \toprule
    Result & Setting & Misclassification & Abstention \\
    \midrule
    \prettyref{thm:main-result} & agnostic & $\sqrt{T^{1+\alpha}\log N}$ & $T^{1-\alpha}+d_\Bc\log^2 T$ \\
    & realizable & $T^{\alpha}\log N$ & $T^{1-\alpha}+d_\Bc\log^2 T$ \\
    \midrule
    \prettyref{thm:censored} (censored) & agnostic & $\sqrt{T^{1+\alpha}\log N}$ & $T^{1-\alpha}+d_\Bc^2\log^4 T$ \\
    & realizable & $T^{\alpha}\log N$ & $T^{1-\alpha}+d_\Bc\log^2 T$ \\
    \bottomrule
  \end{tabular}
  \caption{Summary of our guarantees, for any $\alpha\in[0,1]$ and against any adversary, up to constant factors. In the agnostic setting, we bound $\E[\Mis-\OPT]$ (\prettyref{thm:main-result}) and $\E[\Mis-\OPTc]$ (\prettyref{thm:censored}), and in the realizable setting, we bound $\E[\Mis]$. Here $N=\Nbr{\Hc}{\Bc}{\nicefrac{1}{2T}}$ is the bracketing number of $\Hc$ at scale $\nicefrac{1}{2T}$ using brackets from $\Bc$, and $d_\Bc$ is the VC dimension of the bracket class $\Bc$ (see \prettyref{def:bracket-num} and \prettyref{def:bracket-class}). For halfspaces in $\R^d$, $\log N=O(d\log(dT))$ and $d_\Bc=O(d^2\log d)$ (\prettyref{corr:halfspaces}). For example, $\alpha=\nicefrac13$ gives $T^{2/3}$ rates for both errors in the agnostic setting, and $\alpha=\nicefrac12$ gives $\sqrt{T}$ rates in the realizable setting, up to logarithmic and class-dependent factors.}
  \label{tab:results}
\end{table}

\paragraph{Techniques.} Our algorithmic approach departs significantly from the earlier approaches of \cite{DBLP:conf/nips/GoelHMS23, YB26, EdelmanGoel26}, and is based on the concept of {\em brackets} and {\em bracketing number} (see \prettyref{def:bracket-num} and \prettyref{def:bracket-class}). A brief background and history on bracketing number is deferred to the beginning of \prettyref{sec:refined}. We remark below a few features of our bracketing-based approach.
\begin{itemize}[leftmargin=1.5em]
  \item The agnostic setting presents new technical challenges that do not seem amenable to previous approaches, while our bracketing-based approach overcomes those challenges. A more detailed discussion is presented in \prettyref{sec:technical-overview}.
  \item Earlier results against {\em adaptive} adversaries were known only in the realizable setting and under different structural conditions: finite reduction dimension \citep*{YB26}, and inference dimension and certificate dimension \citep*{EdelmanGoel26}. Our bracketing-based approach provides new techniques to handle adaptive adversaries, which enables us to extend new learning guarantees against adaptive adversaries beyond halfspaces to richer classes such as feed-forward neural networks in the realizable {\em and} agnostic settings.
  \item We highlight that as part of our bracketing-based approach we provide a technique to construct valid brackets from adversarially contaminated sequences. This avoids an {\em exponential} complexity of ``guessing'' the hidden i.i.d.~subsequence \cite{YB26}.
\end{itemize}
\subsection{Related Literature}
\label{sec:related}

\paragraph{Beyond worst-case online learning.}
A broad line of work seeks intermediate models of online learning between the fully-stochastic and fully-adversarial settings. For example, smoothed online learning restricts an adversary by requiring each instance to be generated from a distribution with bounded density ratio relative to a fixed base measure \citep*{DBLP:journals/jacm/HaghtalabRS24,DBLP:conf/colt/BlockDGR22}. Recent work has also removed or weakened the assumption that the base measure is known \citep*{DBLP:conf/colt/BlockRS24, DBLP:conf/stoc/Blanchard25}. More recently, the smoothed online learning framework has been extended to capture more general distributionally-constrained adversaries \citep*{DBLP:journals/corr/abs-2506-10293, DBLP:journals/corr/abs-2602-20585}. A different perspective which keeps the adversary unrestricted but instead considers relaxed benchmarks to compete with has also been studied \citep*{DBLP:conf/colt/MontasserSZ25}. Another classical intermediate model is \emph{transductive} online learning \citep*{BKM97, KK05}, where the adversary fixes the instances $x_1,\dots,x_T$ in advance and reveals them, unlabeled, to the learner, who then predicts their labels online. In the realizable case, the optimal mistake bound for transductive online learning exhibits a trichotomy governed by the VC and Littlestone dimensions \citep*{HMS23}, and its gap to standard online learning was recently tightly quantified \citep*{CHMS25}. One of the main consequences that follow from these different intermediate online learning models is stronger learning guarantees, bypassing the worst-case Littlestone dimension lower bounds and establishing positive guarantees for function classes with bounded VC dimension.

\paragraph{Abstention and selective classification.}
Learning with an abstention option has a long history, beginning with the error--abstention tradeoff of Chow~\cite{Chow70}, and later formalized in several selective classification and learning-with-rejection frameworks (for statistical batch settings) \cite{DBLP:conf/aaai/RivestS88, DBLP:journals/mst/Kivinen95,10.1214/009053604000000058,HerbeiWegkamp06, ElYanivWiener10,CortesDeSalvoMohri16}. For online learning, abstention with fully adversarial sequences has also been studied \cite{DBLP:journals/ml/LiLWS11, DBLP:conf/nips/SayediZB10}, but it is impossible to learn even intervals on the line in this setting. One notable framework that is close to the adversarial injection model considered in this work, is PQ-learning \cite{DBLP:conf/nips/GoldwasserKKM20}. This framework considers a transductive setting where a learning algorithm is presented with labeled training examples from a distribution $P$ and arbitrary unlabeled test examples and is required to classify test examples with low error and low abstention rate under $P$. The adversarial injection model \citep{DBLP:conf/nips/GoelHMS23} can be thought of as the online version of PQ-learning, since the test examples are revealed one-by-one (online) as opposed to all-at-once (transductive). Another closely related model is testable learning with distribution shift (TDS) \cite{DBLP:conf/colt/KlivansSV24}, where the learner is allowed to abstain if the test set is not ``close'' to the training distribution; recent work shows that this coarse-grained model is actually equivalent to PQ-learning \cite{DBLP:journals/corr/abs-2605-07005}. Also related is relatively smart learning \citep*{DP26}, a batch framework in which a learner competes with the best semi-supervised guarantee certifiable from unlabeled data.

\paragraph{Adversarially robust learning.}
The motivation for abstaining on suspicious inputs is related to robustness to adversarial examples \cite{DBLP:journals/corr/SzegedyZSBEGF13, biggio2013evasion, DBLP:journals/corr/GoodfellowSS14}, where the goal is to learn a predictor that is robustly correct on adversarial perturbations of i.i.d.~test examples. There has been a series of works establishing theoretical guarantees for adversarially robust learning \cite[e.g.,][]{DBLP:conf/colt/MontasserHS19, DBLP:conf/alt/AshtianiPU23, DBLP:journals/corr/abs-2306-07723}. Note that this line of work typically assumes ``perturbed'' test examples (e.g., perturbations of bounded norm), while the adversarial injection model studied in this work considers {\em arbitrary} test examples with no restrictions. There are also works focusing on theoretical guarantees against adversarial attacks on the training data \cite[e.g.,][]{DBLP:conf/colt/BlumHQS21, DBLP:conf/colt/BalcanBHS22}.

\section{Preliminaries and Technical Definitions}
\label{sec:prelim}

\paragraph{Problem Setup.}
We denote by $\X$ the instance space, and by $\Hc \subseteq \{0,1\}^{\X}$ the benchmark function class that we would like to compete with. We consider the problem of online classification in the adversarial injection model introduced in \citep*{DBLP:conf/nips/GoelHMS23}. Formally, the learning protocol proceeds as follows. Initially, an adversary (or nature) fixes a distribution $D$ over $\X$. The learner may or may not know the distribution $D$. Then, at each time step $t\in[T]:=\{1,\dots,T\}$:
\begin{enumerate}[leftmargin=2em]
  \item The adversary decides whether or not to inject an adversarial instance by choosing $c_t\in\{1, 0\}$.
  \item If $c_t=0$, then $x_t\iid D$ independently of the past, while if $c_t=1$, then adversary chooses an arbitrary $x_t\in\X$. Afterwards, the adversary chooses an arbitrary label $y_t\in \{0,1\}$ for $x_t$.
  \item The learner observes $x_t$ and outputs $\hat y_t\in\{0,1,\perp\}$, where $\perp$ denotes an abstention.
  \item The learner observes the label $y_t$.
\end{enumerate}
We emphasize that in this work we consider the more general {\em agnostic} setting where the labels $y_t$ chosen by the Adversary are {\em arbitrary}, and need not be consistent with some ground-truth function $h_\star \in \Hc$ as was considered in prior works \citep*{DBLP:conf/nips/GoelHMS23, YB26, EdelmanGoel26}.

If the learner makes a prediction $\hat y_t \in \{0,1\}$ and $\hat y_t \neq y_t$, then this is counted as a misclassification mistake. If the learner abstains $\hat y_t = \perp$ and $x_t$ is iid from $D$ (i.e., $c_t=0$ and there is no adversarial injection), then this is counted as an abstention mistake. Then, the goal for the learner is to make few misclassifications relative to the best-in-class predictor $h_\star\in \Hc$,
\begin{equation}
  \label{eqn:missclassify}
  \Mis - \OPT = \sum_{t=1}^T \1\bigl[\hat y_t\in\{0,1\}\text{ and }\hat y_t\neq y_t\bigr] - \min_{h\in \Hc} \sum_{t=1}^{T} \1[h(x_t) \neq y_t].
\end{equation}
Simultaneously, the learner should make as few abstentions as possible on the iid examples drawn from $D$, i.e., the timesteps with no adversarial injections,
\begin{equation}
  \label{eqn:abstain}
  \Abs:=\sum_{t=1}^T \1\bigl[c_t=0\text{ and }\hat y_t=\perp\bigr].
\end{equation}

\begin{remark}[Censored Variant]
  \label{rem:cens}
  As discussed in \cite{DBLP:conf/nips/GoelHMS23, YB26}, we will also consider a {\em censored} variant of this protocol where the learner observes a feedback label $y_t$ only when it does {\em not} abstain, i.e., $\hat y_t\neq \perp$. In particular, for the agnostic setting we consider in this work, we will additionally require the learner to compete with a {\em stricter} benchmark
  \begin{equation}
    \label{eqn:optc}
    \OPTc := \min_{h \in \Hc} \sum_{t=1}^{T} \1[\hat y_t \neq \perp] \cdot\1[h(x_t)\neq y_t].
  \end{equation}
  Note that $\OPTc \leq \OPT$, since it only considers the best-in-class in a subset of the rounds.
\end{remark}

\begin{remark}[Adaptive vs.~Oblivious]
  \label{rem:censored}
  As discussed in \citep*{YB26}, an oblivious adversary chooses in advance the corruption times $\{t:c_t=1\}$ with their corresponding adversarial injections $(x_t)_{t:c_t=1}$, while an adaptive adversary's decision (and choice) of corruption at time $t$ may depend on the entire history of interaction with the learner before time $t$. The techniques we present will handle both adversaries.
\end{remark}

\paragraph{Brackets.} A central concept and object that will be important for our algorithm and analysis is the notion of {\em brackets} and {\em bracketing number}. For our purposes, a \emph{bracket} is a pair $b=(b^-,b^+)$ of binary functions $b^-,b^+:\X\to\{0,1\}$, such that $b^-(x)\le b^+(x)\text{ for all }x\in\X$. We say that $b$ brackets $h\in\Hc$ if $b^-(x)\le h(x)\le b^+(x)\text{ for every }x\in\X$.

\begin{definition}[Bracketing Number]
  \label{def:bracket-num}
  Let $\Hc$ be a hypothesis class, $D$ be a distribution over $\X$, and let $\eps\in(0,1)$. The $\eps$-\emph{bracketing number} of class $\Hc$ with respect to distribution $D$, denoted by $\Nbr{\Hc}{D}{\eps}$, is the smallest cardinality $N$ of a family of brackets $\mathcal B_{D,\eps} = \{(b_i^-, b_i^+)\}_{i=1}^{N}$ such that for every $h\in\Hc$, there exists $i\in[N]$ satisfying
  \[
    \forall x\in \X: b_i^-(x)\le h(x)\le b_i^+(x)
    \qquad\text{and}\qquad
    \Prob_{X\sim D}(b_i^-(X)\neq b_i^+(X)) \leq \eps.
  \]
  We call such a family an \emph{$\eps$-bracketing} of $\Hc$ with respect to $D$.
\end{definition}

Another concept that will be important for us is the complexity of the bracket class.

\begin{definition} [Complexity of Bracket Class]
  \label{def:bracket-class}
  We say that a class $\mathcal{B} \subseteq \{0,1\}^\X$ brackets $\Hc$ if for every distribution $D$ over $\X$ and every $\eps\in (0,1)$, there exists an $\eps$-bracketing $\mathcal{B}_{D,\eps} \subseteq \Bc\times\Bc$ of $\Hc$ with respect to $D$. Note that this definition requires all binary functions from which brackets for $\Hc$ are constructed to belong to $\Bc$. We denote by $\Nbr{\Hc}{\Bc}{\eps}$ the {\em worst-case} $\eps$-bracketing number of $\Hc$ (using brackets from $\Bc$) over all distributions,
  \begin{equation}
    \Nbr{\Hc}{\Bc}{\eps} := \sup_{D} \inf \left\{|\Bc_{D,\eps}| : \Bc_{D,\eps} \subseteq \Bc\times\Bc \text{ and } \Bc_{D,\eps} \text{ is an $\eps$-bracketing of $\Hc$ wrt $D$} \right\}.
  \end{equation}
  Finally, we denote by $d_\Bc$ the VC dimension of the bracket class $\Bc$.
\end{definition}

\section{Main Results}
\label{sec:mainresult}

Our main result is an algorithm that achieves sublinear misclassification and abstention errors in the {\em agnostic} setting, against oblivious {\em and} adaptive adversaries, for function classes $\Hc$ that admit bounded bracketing number using brackets of bounded VC dimension. Our algorithm is distribution-independent and requires {\em no} prior knowledge of the distribution.

\begin{theorem}\label{thm:main-result}
  Let $\Hc$ be a hypothesis class, and let $\Bc$ be a bracket class for $\Hc$ with VC dimension $d_\Bc$. Let $T\geq 1$ be a time horizon. Then, for any parameter $\alpha \in [0,1]$, \hyperref[alg:agnostic]{Algorithm~\ref*{alg:agnostic}} achieves the following guarantee against {\em any} adversary and any distribution $D$ over $\X$:
  \[
    \E[\Mis - \OPT]
    \lesssim \sqrt{T^{1+\alpha}\log \prn{\Nbr{\Hc}{\Bc}{\nicefrac{1}{2T}}}}~~\text{ and }~~
    \E[\Abs] \lesssim
    T^{1-\alpha}
    +
    d_\Bc \log^{2}\!\left(T\right).
  \]
  Moreover, in the realizable setting where $\OPT=0$, \hyperref[alg:agnostic]{Algorithm~\ref*{alg:agnostic}} satisfies
  \[
    \E[\Mis] \lesssim T^{\alpha} \log \prn{\Nbr{\Hc}{\Bc}{\nicefrac{1}{2T}}} ~~\text{ and }~~
    \E[\Abs] \lesssim
    T^{1-\alpha}
    +
    d_\Bc \log^{2}\!\left(T\right).
  \]
\end{theorem}

\begin{algorithm2e}[H]
  \caption{Online Classification and Abstention via Brackets}
  \label{alg:agnostic}
  \SetKwInput{KwInput}{Input}
  \SetKwInput{KwOutput}{Output}
  \DontPrintSemicolon
  \KwInput{horizon $T$, round length $\tau$, benchmark class $\Hc$, and bracket class $\Bc$ for $\Hc$.}
  Let $R\gets \lceil T/\tau\rceil$.\;
  \For{$r=1 \wedge t=1, \dots, \tau$} {Receive $x_t$ from adversary, output an abstention $\hat y_t = \perp$.}
  \For{$r=2,\dots,R$}{
    Let $I_r= \{(r-1)\tau+1,\dots,\min(r\tau,T)\}$.\;
    Let $S_{r}= (x_1,\dots,x_{(r-1)\tau})$, and $n_{r} = |S_{r}|$.\;
    Let $\mathcal B_r$ be a collection of brackets for $\Hc$ on $S_r$ at scale $\eps = 1/(2n_r)$.\;
    Initialize $p_1$ to be a uniform distribution over the set of experts $\{E_b: b\in\mathcal B_r\}$, each expert $E_b$ is defined as follows
    \[
      E_b(x):= b^-(x) \text{ if } b^-(x)=b^+(x), \text{ otherwise }\perp.
    \]
    \For{$t\in I_r$}{
      Receive $x_t$ from adversary.\;
      Sample a random expert $E_b\sim p_t$ and predict $\hat{y}_t=E_{b}(x_t) \in \{0, 1, \perp\}$.\;
      Upon receiving the feedback label $y_t$, update the mixture $p_t$:
      \[p_{t+1}(E_b) = p_t(E_b)\exp\left( -\eta \1\{E_b(x_t) = 1 - y_t\} \right) / Z_t, \]
      where $Z_t$ is a normalization constant.\;
    }

  }
\end{algorithm2e}

We provide an overview of our algorithmic approach, technical challenges, and proof strategy in \prettyref{sec:technical-overview}. The full proof of \prettyref{thm:main-result} is deferred to \prettyref{sec:main-proof}. Observe that this result exhibits a polynomial tradeoff between misclassification and abstention error rates. Such a polynomial tradeoff is unavoidable even in the realizable case (when $\OPT=0$), as shown in \cite{EdelmanGoel26, YB26}. Specifically, there are function classes with VC dimension 1 such that any algorithm must incur either $\Omega(\sqrt{T})$ misclassification error or $\Omega(\sqrt{T})$ abstention error \citep*{EdelmanGoel26, YB26}.

Next, we present learning guarantees in the {\em censored} variant of the problem (see \prettyref{rem:cens}) where the learner receives label feedback $y_t$ only in the rounds where it does {\em not} abstain ($\hat y_t\neq \perp$). Recall that here we require the learner to compete with a {\em stricter} benchmark, $\OPTc$ in \prettyref{eqn:optc}, which considers the best-in-class only in the rounds where the learner did not abstain.

\begin{theorem}\label{thm:censored}
  Let $\Hc$ be a hypothesis class, and let $\Bc$ be a bracket class for $\Hc$ with VC dimension $d_\Bc$. Let $T\geq 1$ be a time horizon. Then, for any parameter $\alpha \in [0,1]$, \hyperref[alg:censored]{Algorithm~\ref*{alg:censored}} achieves the following guarantee against {\em any} adversary  and any distribution $D$ over $\X$:
  \[
    \E[\Mis - \OPTc]
    \lesssim \sqrt{T^{1+\alpha}\log \prn{\Nbr{\Hc}{\Bc}{\nicefrac{1}{2T}}}} ~~\text{ and }~~
    \E[\Abs] \lesssim
    T^{1-\alpha} + d_\Bc^2 \log^{4}\!\left(T\right).
  \]
  Moreover, in the realizable setting where $\OPTc=0$, \hyperref[alg:censored]{Algorithm~\ref*{alg:censored}} satisfies
  \[
    \E[\Mis] \lesssim T^{\alpha} \log \prn{\Nbr{\Hc}{\Bc}{\nicefrac{1}{2T}}} ~~\text{ and }~~ \E[\Abs] \lesssim
    T^{1-\alpha} + d_\Bc \log^{2}\!\left(T\right).
  \]
\end{theorem}

Note that in the realizable setting, we recover the same guarantee as in the original problem (up to constants), and in the agnostic setting, the only difference is the lower-order term $d_\Bc^2\log^4(T)$, instead of $d_\Bc\log^2(T)$, in the abstention error bound. \prettyref{alg:censored} is a slight variation of \prettyref{alg:agnostic}, and the proof only requires a few changes to the proof of \prettyref{thm:main-result}. We defer those details to \prettyref{sec:censored}.

\begin{remark}[First-Order Rates]
  \label{rem:first-order}
  We remark that because our algorithm is based on Multiplicative Weights, we immediately obtain improved first-order regret bounds of the form
  \[
    \E[\Mis - \OPT] \lesssim \sqrt{T^{\alpha}\,\E[\OPT]\,\log \prn{\Nbr{\Hc}{\Bc}{\nicefrac{1}{2T}}}} + T^{\alpha}\log \prn{\Nbr{\Hc}{\Bc}{\nicefrac{1}{2T}}},
  \]
  while the abstention rate remains the same as in \prettyref{thm:main-result}.
\end{remark}

\subsection{Refined Guarantees for Some Natural Function Classes}
\label{sec:refined}

Given the new generic upper bounds of \prettyref{thm:main-result} in terms of the bracketing number of a class $\Hc$ and the VC dimension of the bracketing class $\Bc$, a natural question to ask here is: which function classes $\Hc$ admit bounded bracketing numbers? At first sight, one might hope that the bracketing number behaves much like the covering number: once a class has finite VC dimension $d$, the $\eps$-covering number scales as $(1/\eps)^{O(d)}$ \cite{Haussler95}. The story, however, is more subtle. While it has been shown that finite VC dimension does indeed imply \emph{finite} $\eps$-bracketing number under every distribution \cite{AdamsNobel10, AdamsNobel12}, there are classes with VC dimension $2$ where the $\eps$-bracketing number can depend arbitrarily badly on $1/\eps$ \cite{Malykhin12, vanHandel13, BKMS21}, ruling out any universal $\eps$-bracketing bound which scales as $(1/\eps)^{O(d)}$. This is not the end of the story. It turns out that for many natural geometric function classes, the $\eps$-bracketing number indeed scales as $(1/\eps)^{O(d)}$. This includes halfspaces (linear classifiers), polynomial threshold functions, and semi-algebraic classes \citep*{BKMS21, DBLP:conf/innovations/DuttaGM22}.

We begin with refined guarantees for the class of halfspaces in $\mathbb{R}^d$, and after that we present a new generic structural result bounding the $\eps$-bracketing number of compositions of classes which enables us to extend our learning guarantees to richer function classes such as intersections of halfspaces and feed-forward neural networks with threshold activations.

\begin{lemma}[Theorem 16 and Lemma 18 of \cite{BKMS21}]\label{lem:halfspaces}
  Let $\Hc = \{x \mapsto \1\{\langle w, x\rangle \leq b\}\mid w \in \mathbb{R}^d, b\in \mathbb{R}\}$ be the class of halfspaces in $\mathbb{R}^d$, and let $\Bc$ be the class of intersections and unions of at most $d+2$ halfspaces from $\Hc$. Then, for any $\eps\in(0,1)$, $\Nbr{\Hc}{\Bc}{\eps}\leq (d/\eps)^{O(d)}$.
\end{lemma}

Combining \prettyref{thm:main-result} with \prettyref{lem:halfspaces} establishes new guarantees for halfspaces in the realizable and agnostic settings as stated in the following corollary.
\begin{corollary}[Halfspaces]\label{corr:halfspaces}
  Let $\Hc$ be halfspaces in $\mathbb{R}^d$. Let $T\geq 1$ be a time horizon. Then, for any parameter $\alpha \in [0,1]$, \hyperref[alg:agnostic]{Algorithm~\ref*{alg:agnostic}} achieves the following guarantee against {\em any} adversary:
  \[
    \E[\Mis - \OPT] \lesssim \sqrt{T^{1+\alpha}\cdot d\log(2dT)} \text{ and }\E[\Abs] \lesssim T^{1-\alpha} + d^{2}\log(d)\log^{2}(T).
  \]
  Moreover, in the realizable setting where $\OPT=0$, \hyperref[alg:agnostic]{Algorithm~\ref*{alg:agnostic}} satisfies
  \[
    \E[\Mis] \lesssim T^{\alpha} \cdot d\log(2dT) ~~\text{ and }~~\E[\Abs] \lesssim T^{1-\alpha} + d^{2}\log(d)\log^{2}(T).
  \]
\end{corollary}
\noindent In particular, for adaptive adversaries in the realizable setting, this result improves on the previous bounds of \cite{YB26} which established an $O(T^{3\alpha})$ misclassification error and an $\Tilde{O}(d^{4.67}T^{1-\alpha})$ abstention error. Observe that \prettyref{corr:halfspaces} immediately implies results for more complex function classes such as $k$-degree polynomial threshold functions in $\mathbb{R}^d$, and more generally any function class with small sign-rank (i.e., embeddable in halfspaces).

Beyond halfspaces and towards establishing learning guarantees for more complex function classes, we present below a new generic structural result bounding the $\eps$-bracketing number of compositions of classes. This result can be viewed as analogous to classical results bounding the VC dimension of compositions of classes \cite{DBLP:journals/jacm/BlumerEHW89, DBLP:journals/theoretics/AlonGHM23}, and extends earlier results bounding the bracketing number of unions and intersections of classes \cite{DBLP:conf/nips/HaghtalabRS20}.

\begin{restatable}[Bracketing Number of Compositions]{lemma}{compositionlemma}\label{lem:composition-bracketing}
  Let $\mathcal H_i\subseteq \{0,1\}^{\mathcal X}$ be classes of binary functions for
  $i\in[k]$, and let $\mathcal F$ be a binary function class over $\{0,1\}^k$ with VC dimension $d_\Fc$.
  Consider the composition of $\Fc$ with $\Hc_1,\dots, \Hc_k$,
  \[
    \mathcal G
    :=
    \left\{
      x\mapsto f\bigl(h_1(x),\dots,h_k(x)\bigr)
      :\ f\in\mathcal F,\ h_i\in\mathcal H_i
    \right\}.
  \]
  Then, for every distribution $D$ over $\X$ and for every $\varepsilon_1,\dots,\varepsilon_k\in[0,1]$,
  \[
    \Nbr{\mathcal G}{D}{\varepsilon_1+\cdots+\varepsilon_k}
    \le
    e\,2^{k d_\Fc}\prod_{i=1}^k \Nbr{\mathcal H_i}{D}{\varepsilon_i}.
  \]
\end{restatable}
\noindent The proof of \prettyref{lem:composition-bracketing} is deferred to \prettyref{app:bracket}. We highlight an interesting aspect of \prettyref{lem:composition-bracketing} which is that the bracketing of composition $\Gc$ depends on the bracketing number of $\Hc_1,\dots, \Hc_k$ and the VC dimension of $\Fc$, and does {\em not} depend on the bracketing number of $\Fc$. Furthermore, as we discuss in detail in \prettyref{sec:technical-overview}, it will be important for our purposes to be able to construct brackets from i.i.d.~samples (that may contain adversarial injections), and we note here that the construction used in the proof of \prettyref{lem:composition-bracketing} preserves this property. That is, if bracketings for $\Hc_i$ can be constructed for (adversarially contaminated) i.i.d.~samples, then a bracketing for $\Gc$ can be constructed from (adversarially contaminated) i.i.d.~samples (see \prettyref{rem:compo-samples}).

Combining \hyperref[lem:halfspaces]{Lemmas~\ref*{lem:halfspaces}} and \ref{lem:composition-bracketing} with \prettyref{thm:main-result} yields new learning guarantees for complex function classes such as intersections of halfspaces and feed-forward neural networks with threshold activations.

\begin{corollary}[Compositions of Halfspaces]
  \label{corr:comp-halfspaces}
  Let $\Gc=\{f(h_1, \dots, h_k): f\in \Fc, h_1,\dots, h_k \in \Hc \}$, where $\Hc$ is the class of halfspaces in $\mathbb{R}^d$ and $\Fc$ is an {\em arbitrary} function class with VC dimension $d_\Fc$. Let $T\geq 1$ be a time horizon. Then, for any parameter $\alpha \in [0,1]$, \hyperref[alg:agnostic]{Algorithm~\ref*{alg:agnostic}} achieves the following guarantee against {\em any} adversary and any distribution $D$ over $\X$:
  \[
    \E[\Mis - \OPT] \lesssim \sqrt{T^{1+\alpha}\cdot k(d_\Fc+d\log(2dkT))} \text{ and }\E[\Abs] \lesssim T^{1-\alpha} + kd^{2}\log(d)\log^{2}(T).
  \]
  Moreover, in the realizable setting where $\OPT=0$, \hyperref[alg:agnostic]{Algorithm~\ref*{alg:agnostic}} satisfies
  \[
    \E[\Mis] \lesssim T^{\alpha} \cdot k(d_\Fc+d\log(2dkT)) ~~\text{ and }~~\E[\Abs] \lesssim T^{1-\alpha} + kd^{2}\log(d)\log^{2}(T).
  \]
\end{corollary}

\paragraph{Guarantees for the censored variation.} Similar to how we established \prettyref{corr:halfspaces} and \prettyref{corr:comp-halfspaces}, we obtain similar quantitative guarantees for the censored variant of the problem (described in \prettyref{rem:censored}) by invoking our generic censored guarantee (\prettyref{thm:censored}) and combining it with \hyperref[lem:halfspaces]{Lemmas~\ref*{lem:halfspaces}} and \ref{lem:composition-bracketing}. The formal statements are deferred to \prettyref{sec:censored}.

\section{Overview of Our Algorithmic Approach}
\label{sec:technical-overview}

\paragraph{Warmup: with prior knowledge of the marginal distribution $D$.} Consider the simpler setting where we know a priori the distribution $D$ over $\X$ generating the i.i.d.~sequence. In this case, our algorithmic approach starts with constructing a finite $\eps$-bracketing $\Bc_{D,\eps}$ for the class $\Hc$ with respect to distribution $D$, where $|\Bc_{D,\eps}|=\Nbr{\Hc}{D}{\eps}$. Given a bracket $b=(b^-,b^+) \in \Bc_{D,\eps}$, we associate with it an abstaining expert
\[
  E_b(x):= b^-(x) \text{ if } b^-(x)=b^+(x), \text{ otherwise }\perp.
\]
By the bracketing definition (\prettyref{def:bracket-num}), observe that for every $h\in \Hc$ there exists a bracket $b=(b^-,b^+)\in \Bc_{D,\eps}$ for $h$ satisfying
\[
  \forall x\in \X: b^-(x) \leq h(x) \leq b^+(x)
  \qquad\text{and}\qquad
  \Prob_{X\sim D}(b^{-}(X)\neq b^+(X)) \leq \eps.
\]
This implies the following {\em key} property: if $b$ brackets $h$, then the abstaining expert $E_b$ never makes more misclassifications than $h$, it either predicts exactly as $h$ or abstains (see \prettyref{lem:dominate}). At the same time, on i.i.d.~samples from $D$, the abstention probability of $E_b$ is exactly the disagreement mass of the bracket, which is at most $\eps$. Therefore, the best expert competes with the best comparator $h\in\Hc$ on misclassification loss, while {\em every} expert abstains rarely on i.i.d.~samples. Thus, running Multiplicative Weights over the finite family of abstaining experts $\{E_b: b\in \Bc_{D,\eps}\}$ would guarantee
\[
  \E[\Mis - \OPT] \lesssim \sqrt{T\cdot \log \Nbr{\Hc}{D}{\eps}}
  \quad\text{and}\quad
  \E[\Abs]\leq \sum_{t=1}^{T} \Prob(\hat y_t = \perp | c_t=0) \leq \eps T.
\]
For completeness, we present a formal claim and proof of this simpler setting in \prettyref{app:known-d-baseline}.

\paragraph{Challenge: with {\em no} prior knowledge of marginal distribution $D$.} The actual problem is harder because the learning algorithm does {\em not} know the distribution $D$, and moreover does not know which observed examples are i.i.d.~draws from $D$ and which are adversarial injections. Thus, we cannot simply precompute a bracketing $\Bc_{D,\eps}$, nor can we first identify the i.i.d.~subsequence and then construct brackets based on this subsequence. The central challenge is therefore:
\begin{center}
  \textit{How to construct valid brackets directly from sequences that are contaminated with adversarial injections, without first identifying which points are iid?}
\end{center}
We first discuss a tempting strategy that fails in the agnostic setting. In the realizable setting, Yu and Blanchard~\cite{YB26} provide a boosting framework which runs exponentially many weak learners on ``guessed'' i.i.d.~subsequences. Each weak learner treats its given subsequence as i.i.d.~and proceeds to run a sample-based online learning algorithm for prediction with abstention. The booster guarantees an abstention error close to that of the best weak learner up to an additive term, but pays a multiplicative overhead in misclassification error. A natural adaptation here is therefore to build bracket-experts from every candidate i.i.d.~subsequence of the observed history and feed them to the same booster. The difficulty is that our goal in the agnostic setting is to compete with the best-in-class ($\OPT$) up to an \emph{additive} regret term, whereas the boosting strategy of Yu and Blanchard~\cite{YB26} would incur a prohibitively large multiplicative factor in front of $\OPT$.

We take a different approach that avoids the complexity of guessing exponentially many possibilities. Specifically, our algorithm does {\em not} need to guess i.i.d.~subsequences of the history, and instead directly constructs brackets from the entire history, even though it may contain adversarial injections. Recall that \prettyref{alg:agnostic} has access to a class $\Bc$ that brackets $\Hc$ (as in \prettyref{def:bracket-class}). It partitions the time horizon into blocks of length $\tau$, and at the beginning of each block $r$ it uses the entire (possibly contaminated) history $S_r$ to construct a fresh bracketing family $\mathcal B_r$ for $\Hc$ at empirical scale $\eps=1/(2n_r)$, where $n_r=|S_r|$. It then runs a new Multiplicative Weights instance over the induced abstaining experts $\{E_b:b\in \mathcal B_r\}$ during this block.

The misclassification analysis is then relatively standard: on each block $r$, there is a bracket $b\in\mathcal B_r$ which brackets the best-in-class comparator $h^\star$, and so the corresponding expert $E_b$ either abstains or predicts exactly as $h^\star$ (see \prettyref{lem:dominate}). Thus,
invoking the regret guarantee of Multiplicative Weights (\hyperref[lem:multiplicativeweights]{Lemma~\ref*{lem:multiplicativeweights}}) on each block and summing over blocks yields
\[
  \E[\Mis-\OPT]
  \lesssim
  \sum_{r=2}^R\sqrt{\tau \log |\mathcal B_r|} \leq \sqrt{\frac{T^2}{\tau}\log \Nbr{\Hc}{\Bc}{\nicefrac{1}{2T}}}.
\]
The more challenging part is the abstention analysis. How can brackets constructed from i.i.d.~samples have low disagreement mass? And, how can this carry through under adversarial injections? Recall here that all brackets $b=(b^{-}, b^{+})$ belong to a class $\Bc$ of finite VC dimension $d_\Bc$ (\prettyref{def:bracket-class}). It follows that the disagreement class $\{x\mapsto \1[b(x)\neq b'(x)]: b,b'\in\Bc\}$ has finite VC dimension $O(d_\Bc)$ \cite{DBLP:journals/jacm/BlumerEHW89}. Therefore, with enough i.i.d.~samples, any bracket $b=(b^-, b^+)$ with low disagreement on the empirical sample will have low disagreement on the underlying distribution. In \hyperref[thm:onesided]{Lemma~\ref*{thm:onesided}}, we appeal to standard uniform convergence bounds to argue that with high probability over the draw of $m$ iid samples from $D$,
\[
  \forall b^-,b^+\in\Bc,~~\frac{1}{m} \sum_{i=1}^{m} \1\{b^-(x)\neq b^+(x)\} = 0 \qquad\Longrightarrow\qquad \Prob_D(b^-(X)\neq b^+(X)) \lesssim \frac{d_\Bc\log(m)}{m}.
\]
In a \emph{key} \hyperref[lem:contam]{Lemma~\ref*{lem:contam}}, we prove that this remains true even when the adversary, after seeing the $m$ iid samples, decides to inject additional $n-m$ adversarial points. Specifically, we show that if the empirical disagreement of a bracket $b=(b^-, b^+)\in \Bc_r$ on the full observed sequence $S_r$ is at most $1/(2n_r)$, then in fact the disagreement is zero on the i.i.d.~subsequence hidden inside $S_r$, and therefore the probability that an expert $E_b$ abstains under distribution $D$ is small.

Given that each block $r$ in \prettyref{alg:agnostic} has length $\tau$, there can be at most $2\tau$ abstentions on i.i.d.~instances before \prettyref{alg:agnostic} collects at least $\tau$ i.i.d.~instances. After that, for blocks $r$ in which the number of i.i.d.~instances $m_r$ available in $S_r$ is at least $\tau$, every bracket in $\mathcal B_r$ abstains on a fresh clean example with probability $\Tilde{O}(d_\Bc/m_r)$, which sums to only a logarithmic contribution over later blocks. Choosing $\tau=\ceil{T^{1-\alpha}}$ yields the tradeoff of \prettyref{thm:main-result}. The full formal proof is deferred to \prettyref{sec:main-proof}. Notably, \prettyref{alg:agnostic} need not detect when its brackets have low abstention probability: more i.i.d.~instances simply yield better brackets, while more adversarial injections can be safely handled by abstaining without penalty.

\section{\texorpdfstring{Proof of Theorem~\ref*{thm:main-result}}{Proof of Theorem 1}}
\label{sec:main-proof}

In this section, we provide a full proof of \prettyref{thm:main-result}. We split the proof into two main parts: (a) misclassification error analysis, and (b) abstention error analysis.

\subsection{Misclassification Error Analysis}

We begin with stating a standard Lemma on the regret guarantee of Multiplicative Weights.

\begin{lemma} [See, e.g.~Theorem 2.2 and Theorem 2.4 in \cite{cesa2006prediction}]
  \label{lem:multiplicativeweights}
  Given a finite set of experts $F=\{f_1,\dots, f_N\}$ and an arbitrary sequence of loss functions $\ell_1,\dots, \ell_T: F \to [0,1]$, running the Multiplicative Weights algorithm using experts $F$ with parameter $\eta>0$ produces a sequence of mixture strategies $p_1,\dots, p_T$ which satisfy
  \[
    \sum_{t=1}^{T} \E_{f\sim p_t} \ell_t(f) - L^\star \leq \frac{\log N}{\eta} + \frac{\eta T}{8}, \quad \text{ where }L^\star:=\min_{f\in F}\sum_{t=1}^{T}\ell_t(f).
  \]
  In particular, choosing $\eta = \sqrt{8\log N/T}$ guarantees $\sum_{t=1}^{T} \E_{f\sim p_t} \ell_t(f) - L^\star \leq \sqrt{(T/2)\log N}$. Moreover, if $L^\star=0$, choosing $\eta \geq \log(2)$, guarantees $\sum_{t=1}^{T} \E_{f\sim p_t} \ell_t(f) \leq 2 \log N$. These guarantees hold for every sequence of loss functions, including sequences chosen adaptively (e.g., by an adaptive adversary), since each $p_t$ is a deterministic function of $\ell_1,\dots,\ell_{t-1}$.
\end{lemma}

The following {\em key} Lemma states that an expert $E_b$ based on a bracket $b$ for a predictor $h$ never makes more misclassifications than $h$.

\begin{lemma}\label{lem:dominate}
  Let $h\in\Hc$, let $b=(b^-,b^+)$ bracket $h$, and let $(x,y)$ be any labeled example.
  Then
  \[
    \1\bigl[E_b(x)\in\{0,1\}\text{ and }E_b(x)\neq y\bigr]
    \le
    \1\bigl[h(x)\neq y\bigr].
  \]
\end{lemma}

\begin{proof}
  If $E_b(x)=1$, then $b^-(x)=1$, hence $h(x)=1$ because $b^-\le h$.
  Thus the left-hand side equals $\1[y=0]=\1[h(x)\neq y]$.
  If $E_b(x)=0$, then $b^+(x)=0$, hence $h(x)=0$ because $h\le b^+$.
  Thus the left-hand side equals $\1[y=1]=\1[h(x)\neq y]$.
  If $E_b(x)=\perp$, the left-hand side is $0$.
\end{proof}

We are now ready to bound the misclassification error.

\begin{lemma}[Misclassification Error]\label{lem:mis-generic}
  For any distribution $D$ over $\X$ and any (possibly adaptive) adversary, \hyperref[alg:agnostic]{Algorithm~\ref*{alg:agnostic}} satisfies
  \[
    \E\brk{\Mis - \OPT}
    \le
    \sqrt{(T^2/2\tau) \cdot \log \Nbr{\Hc}{\Bc}{\nicefrac{1}{2T}}}.
  \]
  Moreover, in the realizable case where $\OPT=0$,
  \[
    \E\brk{\Mis}\le 2(T/\tau) \cdot \log \Nbr{\Hc}{\Bc}{\nicefrac{1}{2T}}.
  \]
\end{lemma}

\begin{proof}
  Recall that \prettyref{alg:agnostic} divides the time horizon $\set{1,\dots, T}$ into blocks of length $\tau$, with a total of $R=\ceil{T/\tau}$ blocks. For each block $r\in\{1,\dots,R\}$, recall that $I_r=\{(r-1)\tau+1,\dots,\min(r\tau,T)\}$ is the set of timesteps in block $r$.

  For each timestep $t$, let $\mathsf H_{t-1}$ denote the history of the interaction before time $t$ (including the adversary's choices and \prettyref{alg:agnostic}'s random choices and predictions), and let $\mathsf H^{+}_{t-1}$ denote $\mathsf H_{t-1}$ together with $(c_t,x_t,y_t)$, i.e., the history after the adversary has selected the current example and its label, but before \prettyref{alg:agnostic} randomizes at time $t$. For each block $r\in\{2,\dots,R\}$, timestep $t\in I_r$, and bracket $b\in\mathcal B_r$, let $\ell_t(E_b):=\1\set{E_b(x_t)=1-y_t}$ denote the misclassification loss of expert $E_b$ at time $t$. Observe that the family $\mathcal B_r$ and the mixture $p_t$ are determined by $\mathsf H_{t-1}$, and that \prettyref{alg:agnostic} samples $E_b\sim p_t$ using fresh randomness that is independent of $\mathsf H^{+}_{t-1}$. Therefore,
  \[
    \E\left[\1\bigl[\hat y_t\in\{0,1\}\text{ and }\hat y_t\neq y_t\bigr] \,\middle|\, \mathsf H^{+}_{t-1}\right] = \E_{E_b\sim p_t}\,\ell_t(E_b).
  \]
  Since \prettyref{alg:agnostic} abstains on every timestep of block $1$, by the law of total expectation,
  \[
    \E\brk{\Mis} = \E\left[\sum_{r=2}^{R}\sum_{t\in I_r}\E_{E_b\sim p_t}\,\ell_t(E_b)\right].
  \]
  We will bound the quantity inside the expectation for every realization of the interaction between the adversary and \prettyref{alg:agnostic}.

  Fix such a realization, and let $h_\star\in\arg\min_{h\in\Hc}\sum_{t=1}^T \1\bigl[h(x_t)\neq y_t\bigr]$ for the realized sequence $(x_t,y_t)_{t=1}^T$. For each  block $r\in\{2,\dots,R\}$, the family $\mathcal B_r$ contains some bracket $b_{r,\star} = (b_{r,\star}^{-}, b_{r,\star}^{+})$ such that
  \[
    b_{r,\star}^{-}(x)\le h_\star(x)\le b^+_{r,\star}(x) ~~\forall x\in \X.
  \]
  By invoking \hyperref[lem:dominate]{Lemma~\ref*{lem:dominate}}, it follows that the total misclassification error of expert $E_{b_{r, \star}}$ is bounded from above by the total misclassification error of $h_\star$,
  \[
    \sum_{t\in I_r} \ell_t(E_{b_{r, \star}}) \le \sum_{t\in I_r}\1\bigl[h_\star(x_t)\neq y_t\bigr].
  \]
  Within block $r$, the mixtures $(p_t)_{t\in I_r}$ are the iterates of Multiplicative Weights on the loss functions $(\ell_t)_{t\in I_r}$, and the regret guarantee of Multiplicative Weights (\hyperref[lem:multiplicativeweights]{Lemma~\ref*{lem:multiplicativeweights}}) holds for every sequence of loss functions, in particular for the realized one, even though it may have been chosen adaptively. Combining this with the inequality above, and recalling that $|I_r|\leq\tau$, gives
  \[
    \sum_{t\in I_r}\E_{E_b\sim p_t}\,\ell_t(E_b) \leq \sum_{t\in I_r}\1\bigl[h_\star(x_t)\neq y_t\bigr] + \sqrt{\frac{\tau}{2}\log|\Bc_r|}.
  \]
  Summing over $r=2,\dots,R$ yields
  \begin{align*}
    \sum_{r=2}^{R}\sum_{t\in I_r}\E_{E_b\sim p_t}\,\ell_t(E_b)
    &\le
    \sum_{r=2}^{R} \sum_{t\in I_r}\1\bigl[h_\star(x_t)\neq y_t\bigr] + \sum_{r=2}^R \sqrt{\frac{\tau}{2}\log|\Bc_r|}\\
    &\leq \OPT +  \sqrt{\frac{T^2}{2\tau}\log \Nbr{\Hc}{\Bc}{\nicefrac{1}{2T}}}.
  \end{align*}
  Since this holds for every realization of the interaction, taking expectations proves the first inequality. The proof of the second inequality is identical, but we invoke the realizable guarantee of Multiplicative Weights instead (\hyperref[lem:multiplicativeweights]{Lemma~\ref*{lem:multiplicativeweights}}), which applies since $\OPT=0$ implies that $\sum_{t\in I_r}\ell_t(E_{b_{r,\star}})=0$ for every block $r$.
\end{proof}

\subsection{Abstention Error Analysis}
We now turn to bounding the abstention error of \prettyref{alg:agnostic}. We begin with stating a standard uniform convergence guarantee for VC classes.

\begin{lemma}[Theorem A2.1 and Proposition A2.1 of \cite{DBLP:journals/jacm/BlumerEHW89}]\label{thm:onesided}
  Let $\Gc \subseteq \set{0,1}^\X$ be a binary class with finite VC dimension $d_\Gc<\infty$. Then, for any $\delta \in (0,1), m \geq 1$, and any distribution $D$ over $\X$, with probability at least $1-\delta$ over $X_1, \dots, X_m \iid D$,
  \[
    \forall g\in\Gc:~~ \frac{1}{m}\sum_{i=1}^{m} g(X_i) = 0 \qquad\Longrightarrow\qquad\Prob_D(g(X)=1) \leq \frac2m \left( d_\Gc \log\left(\frac{2em}{d_\Gc}\right) + \log\left(\frac{2}{\delta}\right)\right).
  \]
\end{lemma}

We are now ready to state and prove a {\em key} Lemma showing that if we are constructing brackets for $\Hc$ using a bracketing class $\Bc$ with bounded VC dimension (see \prettyref{def:bracket-class}), then it is possible to construct such brackets from i.i.d.~samples from a distribution $D$ even after an adversary contaminates the samples with arbitrary adversarial injections. Note that the adversarial injections may be chosen \emph{after seeing the entire} i.i.d.~{\em sample}, and they may depend arbitrarily on it.

\begin{lemma}[Constructing Brackets with Adversarially Contaminated Samples]\label{lem:contam}
  Fix a bracketing class $\Bc \subseteq \set{0,1}^\X$, a distribution $D$ on $\X$, an integer $m\geq 1$ and $\delta \in (0,1)$. With probability at least $1-\delta$ over $Z_1,\dots, Z_m \iid D$:

  \indent for every $n\geq m$, every sequence $S=(x_1,\dots, x_n)$ which contains $Z_1,\dots, Z_m$, and every $b^-, b^+\in \Bc$,
  \[
    \frac{1}{n}\sum_{i=1}^{n}  \1\{b^-(x_i) \neq b^+(x_i)\} \leq \frac{1}{2n} \Longrightarrow \Prob_D(b^-(Z) \neq b^+(Z)) \leq \frac{2}{m}\left( d_\Bc \log\left(\frac{2em}{d_\Bc}\right) + \log\left(\frac{2}{\delta}\right)\right).
  \]
\end{lemma}

\begin{proof}
  Consider the class of parities over $\Bc$: $\set{x \mapsto \1\set{b(x) \neq b'(x)}\mid b, b ' \in \Bc}$. By classical results \cite{DBLP:journals/jacm/BlumerEHW89}, it follows that the VC dimension of this class is at most $O(d_\Bc)$. Let $A$ denote the event over $Z_1,\dots, Z_m \iid D$ that for every $ b^-, b^+ \in \Bc$,
  \[\frac{1}{m}\sum_{i=1}^{m}  \1\{b^-(Z_i) \neq b^+(Z_i)\} = 0 \Longrightarrow \Prob_D(b^-(Z) \neq b^+(Z)) \leq \frac{2}{m}\left( d_\Bc \log\left(\frac{2em}{d_\Bc}\right) + \log\left(\frac{2}{\delta}\right)\right).\]
  By invoking \hyperref[thm:onesided]{Lemma~\ref*{thm:onesided}}, it follows that $\Prob(A) \geq 1-\delta$. Observe now that for any $Z_1,\dots, Z_m \iid D$ and any sequence $S\in \X^n$ which contains $Z_1,\dots, Z_m$, for every $ b^-, b^+ \in \Bc$,
  \begin{align*}
    \frac{1}{n}\sum_{x\in S}  \1\{b^-(x) \neq b^+(x)\} \leq \frac{1}{2n}
    &\Longrightarrow \frac{1}{n}\sum_{x\in S} \1\{b^-(x) \neq b^+(x)\} = 0\\
    &\Longrightarrow \frac{1}{m}\sum_{i=1}^{m} \1\{b^-(Z_i) \neq b^+(Z_i)\} = 0.
  \end{align*}
  Therefore, it follows that event $A$ implies the target event $A'$ over $Z_1,\dots, Z_m \iid D$:
  \begin{align*}
    \forall &n\geq m, \forall S\in\X^n \text{ such that } \forall_{1 \leq i\leq m} Z_i \in S, \forall b^-, b^+\in \Bc,\\
    &\frac{1}{n}\sum_{x\in S} \1\{b^-(x) \neq b^+(x)\} \leq \frac{1}{2n} \Longrightarrow \Prob_D(b^-(Z) \neq b^+(Z)) \leq \frac{2}{m}\left( d_\Bc \log\left(\frac{2em}{d_\Bc}\right) + \log\left(\frac{2}{\delta}\right)\right).
  \end{align*}
  Thus, $\Prob(A') \geq 1-\delta$.
  This argument is deterministic once the clean sample is fixed, and therefore it holds simultaneously for all realized contamination sequences, regardless of how they were selected based on $(Z_1,\dots,Z_m)$.
\end{proof}

We are now ready to proceed with bounding the abstention error of \prettyref{alg:agnostic}.

\begin{lemma}[Abstention Error]\label{lem:abstentions}
  \hyperref[alg:agnostic]{Algorithm~\ref*{alg:agnostic}} satisfies
  \[
    \E[\Abs]
    \le
    2\tau
    +
    O\left( d_\Bc \log(T)\right)\log\prn{\frac{T}{\tau}}.
  \]
\end{lemma}

\begin{proof}
  Recall that \prettyref{alg:agnostic} divides the time horizon $\set{1,\dots, T}$ into blocks of length $\tau$, with a total of $R=\ceil{T/\tau}$ blocks. For each block $r\in\{1,\dots,R\}$, recalling that $I_r=\{(r-1)\tau+1,\dots,\min(r\tau,T)\}$ is the set of timesteps in block $r$, let
  \[
    m_r:=\sum_{t<(r-1)\tau+1}\1[c_t=0],
    \qquad
    M_r:=\sum_{t\in I_r}\1[c_t=0].
  \]
  So $m_r$ is the number of clean iid examples before the start of block $r$, and $M_r$ is the number of clean iid examples appearing during block $r$.
  \paragraph{An equivalent view of interaction.} We first describe a helpful way of thinking about the interaction between \prettyref{alg:agnostic} and the adversary. Let $Z_1,Z_2,\ldots\iid D$ be an auxiliary sequence generated in advance independently of the adversary. At time $t$, let $N_{t-1}:=\sum_{s<t}\1\{c_s=0\}$. If the adversary chooses $c_t=0$, set $x_t:=Z_{N_{t-1}+1}$; if it chooses $c_t=1$, then the adversary chooses an arbitrary adversarial injection $x_t\in \X$. Because the adversary decides $c_t\in\set{0,1}$ before seeing $Z_{N_{t-1}+1}$, it holds that $Z_{N_{t-1}+1}\sim D$ and is independent of the past, so this construction has the same law as the original interaction. Moreover, almost surely, for every $t$, the clean examples among $x_1,\dots,x_t$, listed chronologically, are exactly $Z_1,\dots,Z_{N_t}$. In particular, for every block $r$, the history $S_r$ contains $Z_1,\dots,Z_{m_r}$ as a subsequence. It therefore suffices to prove the claimed bound under this view.

  We now proceed with the analysis of the abstention error, noting that throughout the proof, expectations are over the draws of iid examples $Z_1, Z_2, \dots, \iid D$ (as in the view described above), the adversary's possible randomness, and \prettyref{alg:agnostic}'s internal randomness.

  \noindent Let $r_\tau$ be the first block where the number of iid examples available before the start of the block is at least $\tau$
  \[
    r_\tau:=\min\{r\in\{2,\dots,R+1\}: m_r\ge \tau\},
  \]
  with the convention $r_\tau=R+1$ if no such block exists.

  We will analyze the expected $\Abs$ on iid examples by splitting it into two parts: (a) expected $\Abs$ for all blocks $r < r_\tau$ (before collecting $\tau$ iid examples), and (b) expected $\Abs$ for all blocks $r \geq r_\tau$ (after collecting at least $\tau$ iid examples).

  \begin{align}
    \E\left[ \Abs \right] & = \E\left[ \sum_{r=1}^{R} \sum_{t\in I_r} \1\{c_t=0~\wedge~\hat{y}_t=\perp\}\right]                                                                            \\
    & = \E \left[ \sum_{r < r_\tau}\sum_{t\in I_r} \1\{c_t=0~\wedge~\hat{y}_t=\perp\} + \sum_{r\geq r_\tau}\sum_{t\in I_r} \1\{c_t=0~\wedge~\hat{y}_t=\perp\}\right].
  \end{align}

  \paragraph{Abstention error before collecting $\tau$ iid examples.}  The total number of clean iid examples in blocks $1,\dots,r_\tau-1$ equals $m_{r_\tau}$. If $r_\tau=R+1$, then $m_{r_\tau}<\tau$ by definition.
  Otherwise, by definition of $r_\tau$, note that $m_{r_\tau-1}<\tau$, and because each block has length $\tau$, the number of iid examples that appeared in block $r_\tau - 1$ can be at most $\tau$, $M_{r_\tau-1}\le \tau$. It follows then that $m_{r_\tau}=m_{r_\tau-1}+M_{r_\tau-1}<2\tau$. Therefore, the maximum number of abstentions on clean iid examples that happen in blocks $1,\dots, r_\tau - 1$ is at most $2\tau-1$:
  \[\E \left[ \sum_{r < r_\tau}\sum_{t\in I_r} \1\{c_t=0~\wedge~\hat{y}_t=\perp\}\right] \leq 2\tau-1.\]

  \paragraph{Abstention error after collecting at least $\tau$ iid examples.} For the remainder of the analysis, we will need to introduce probability events that capture whether the (random) iid examples collected can be used to construct valid brackets with low abstention rate. Specifically, for each $j\in\{1,\dots,T\}$, let $A_j$ be the event in \hyperref[lem:contam]{Lemma~\ref*{lem:contam}} for the clean iid examples $Z_1,\dots,Z_j$ with confidence parameter $\delta_j= 1/T^2$. On event $A_j$, every adversarially contaminated sequence $S\in \X^n$ containing the clean examples $Z_1,\dots, Z_j$ and every $b^-, b^+ \in \Bc$ satisfy
  \begin{equation}
    \label{eqn:contam}
    \frac{1}{n}\sum_{x\in S} \1\set{b^-(x)\neq b^+(x)} \leq \frac{1}{2n}\quad\Longrightarrow\quad\Prob_D(b^-(Z)\neq b^+(Z)) \leq \frac{O\left( d_\Bc \log(T)\right)}{j}.
  \end{equation}

  Now, recall that in each round $r\in \{2, \dots, R\}$, \hyperref[alg:agnostic]{Algorithm~\ref*{alg:agnostic}} works by constructing brackets $\mathcal{B}_r$ based on the entire sequence of examples $S_r$ available so far. However, we do not know a priori the number of clean examples $m_r\in \{1,\dots, T\}$ that are contained in $S_r$ since it depends on the interaction between the adversary and \hyperref[alg:agnostic]{Algorithm~\ref*{alg:agnostic}} before the start of block $r$. So, we do not know apriori which events $A_j$ for $j\in\{1,\dots, T\}$ will be relevant for the analysis of \hyperref[alg:agnostic]{Algorithm~\ref*{alg:agnostic}}. Hence, we will consider the intersection event
  \[A := \bigcap_{j=1}^{T}A_j,\]
  where all events $A_1,\dots, A_T$ hold simultaneously. By a union bound,
  \[ \Prob( \bar{A} ) \leq \sum_{j=1}^{T}\Prob(\bar{A}_j)\leq \sum_{j=1}^{T} \frac{1}{T^2} = \frac{1}{T}.\]

  When event $A$ does not hold, the total abstention error on clean examples is at most $T$, therefore, by law of total expectation,
  \begin{align}
    \E\left[ \sum_{r\geq r_\tau}\sum_{t\in I_r} \1\{c_t=0~\wedge~\hat{y}_t=\perp\} \right] & \leq \E\left[ \1\{A\}\sum_{r\geq r_\tau}\sum_{t\in I_r} \1\{c_t=0~\wedge~\hat{y}_t=\perp\} \right] + \Prob(\bar{A}) T \\
    & \leq \E\left[ \1\{A\}\sum_{r\geq r_\tau}\sum_{t\in I_r} \1\{c_t=0~\wedge~\hat{y}_t=\perp\} \right] + 1.
  \end{align}
  Thus, it remains to complete the analysis of abstention error when event $A$ holds. Fix a block $r\in\{ r_\tau, \dots, R\}$. By definition of $r_\tau$, the number of iid examples available before the start of block $r$, $m_r \geq \tau$. Note that the set of collected examples so far $S_r$ contains $Z_1,\dots, Z_{m_r}$ as a subset. By the construction of the brackets $\mathcal{B}_r$, every bracket $b=(b^-, b^+)\in \mathcal{B}_r$ satisfies
  \[
    \frac{1}{n_r}\sum_{x\in S_r} \1\set{b^-(x)\neq b^+(x)} \leq \frac{1}{2n_r}.
  \]
  Therefore, under event $A_{m_r}$, \prettyref{eqn:contam} implies
  \[
    \Prob_{Z\sim D}(E_b(Z) = \perp ) \leq \frac{O\left( d_\Bc \log(T)\right)}{m_r}\quad \text{ for every } b \in \mathcal{B}_r.
  \]
  So, we can now use this property to argue that the expected number of abstentions on iid examples in block $r$ will be low because all experts have a low probability of abstaining.

  Fix a timestep $t\in I_r$ in block $r$, and condition on the history of the interaction up to time $t-1$. Observe that the mixture $p_t$ over experts in $\mathcal{B}_r$ maintained by \prettyref{alg:agnostic} is already fixed. If $c_t=0$, then $x_t=Z_{N_{t-1}+1}$ and $Z_{N_{t-1}+1}\sim D$ independent of the history so far, thus conditioning on $A_{m_r}$,
  \[
    \Prob\left(\hat{y}_t = \perp | c_t=0, A_{m_r} \right) = \sum_{b\in \mathcal{B}_r} p_t(E_b) \Prob(E_b=\perp) \leq \sum_{b\in \mathcal{B}_r} \prn{p_t(E_b)  \frac{O\left( d_\Bc \log(T)\right)}{m_r}} =  \frac{O\left( d_\Bc \log(T)\right)}{m_r}.
  \]
  Summing over $t\in I_r$ gives
  \[\E\left[ \1\{A_{m_r}\}\sum_{t\in I_r} \1\{c_t=0~\wedge~\hat{y}_t=\perp\}\right] \leq \E\left[M_r \frac{O\left( d_\Bc \log(T)\right)}{m_r} \right].\]
  Since $A\subseteq A_{m_r}$, summing over all blocks $r \geq r_\tau$ yields
  \[
    \E\left[ \1\{A\}\sum_{r=r_\tau}^{R}\sum_{t\in I_r} \1\{c_t=0~\wedge~\hat{y}_t=\perp\}\right] \leq \E\left[\sum_{r=r_\tau}^{R}\frac{M_r}{m_r} \right]\cdot O\left( d_\Bc \log(T)\right).
  \]
  For every late block $r\ge r_\tau$, we have $m_r\ge \tau$ and $M_r\le \tau$, so $0 \leq M_r/m_r \leq 1$. Using $u \leq 2\log(1+u)$ for $u\in [0,1]$ and $m_{r+1} = m_r + M_r$,
  \[
    \frac{M_r}{m_r} \leq 2\log\left(1+\frac{M_r}{m_r}\right) = 2\log\left(\frac{m_{r+1}}{m_r}\right).
  \]
  Consequently, summing over $r=r_\tau,\dots,R$ telescopes to
  \[
    \sum_{r=r_\tau}^R \frac{M_r}{m_r}
    \le
    2\sum_{r=r_\tau}^R \log\prn{\frac{m_{r+1}}{m_r}}
    \le
    2\log\prn{\frac{m_{R+1}}{m_{r_\tau}}}
    \le
    2\log\prn{\frac{T}{\tau}}.
  \]
  Putting everything together,
  \begin{align}
    \E\left[ \Abs \right] & = \E\left[ \sum_{r=1}^{R} \sum_{t\in I_r} \1\{c_t=0~\wedge~\hat{y}_t=\perp\}\right]                            \\
    & \leq  2\tau-1 + \E\left[ \1\{A\}\sum_{r=r_\tau}^{R}\sum_{t\in I_r} \1\{c_t=0~\wedge~\hat{y}_t=\perp\}\right] + 1 \\
    & \leq 2\tau + O\left( d_\Bc \log(T)\right)\log\prn{\frac{T}{\tau}}.
  \end{align}
  This concludes the proof.
\end{proof}

\begin{proof}[Proof of {\prettyref{thm:main-result}}]
  The proof follows from \prettyref{lem:mis-generic} and \prettyref{lem:abstentions}, by choosing $\tau=\ceil{T^{1-\alpha}}$.
\end{proof}

\section{Censored Feedback and Competing with a Refined OPT}
\label{sec:censored}

In this section, we focus on a variant of the problem where the learner is not provided with feedback labels $y_t$ in the rounds or timesteps it chooses to abstain $\hat y_t=\perp$ (see \prettyref{rem:cens}). Importantly, in the agnostic setting, we will require the learner to compete with the best-in-class only in the rounds where the learner did not abstain, which we recall here
\begin{equation*}
  \OPTc = \min_{h \in \Hc} \sum_{t=1}^{T} \1[\hat y_t \neq \perp] \cdot\1[h(x_t)\neq y_t].
\end{equation*}
Note that $\OPTc$ is a more refined benchmark than $\OPT$ since $\OPTc \leq \OPT$. Throughout this section, we write $N:=\Nbr{\Hc}{\Bc}{\nicefrac{1}{2T}}$, and assume that $N\geq 3$, so that $\log N\geq 1$. We begin by restating \prettyref{thm:censored}.

\begin{restatedtheorem}{Theorem~\ref*{thm:censored}}[Restated]
  Let $\Hc$ be a hypothesis class, and let $\Bc$ be a bracket class for $\Hc$ with VC dimension $d_\Bc$. Let $T\geq 1$ be a time horizon. Then, for any parameter $\alpha \in [0,1]$, \prettyref{alg:censored} achieves the following guarantee against {\em any} adversary and any distribution $D$ over $\X$:
  \[
    \E[\Mis - \OPTc] \leq 4\sqrt{T^{1+\alpha}\log N}
    \quad\text{and}\quad
    \E[\Abs] \leq 3T^{1-\alpha} + O\prn{d_\Bc^2\log^{4} T}.
  \]
  Moreover, in the realizable setting where $\OPTc=0$, \prettyref{alg:censored} satisfies
  \[
    \E[\Mis] \leq 4 T^{\alpha} \log N
    \quad\text{and}\quad
    \E[\Abs] \leq 2T^{1-\alpha} + O\prn{d_\Bc \log^{2} T}.
  \]
\end{restatedtheorem}

\paragraph{Algorithm.} \prettyref{alg:censored} has the same block structure and uses the same bracketing families $\mathcal B_r$ and experts $\set{E_b: b\in\mathcal B_r}$ as \prettyref{alg:agnostic}. It only replaces the inner loop over the timesteps $t\in I_r$ of each block $r\in\set{2,\dots,R}$. At each timestep, it computes the mass $a_t$ of the experts that abstain on $x_t$. If $a_t>\rho$, it abstains and therefore receives no label. Otherwise, it predicts using an expert sampled from the experts that do not abstain on $x_t$, and updates its mixture as in \prettyref{alg:agnostic}. The realizable and agnostic settings only differ in the choice of the parameters $\rho$ and $\eta$ (see the proof of \prettyref{thm:censored} below).

\begin{algorithm2e}[H]
  \caption{Censored variant of Algorithm~\ref{alg:agnostic} (inner loop of block $r$)}
  \label{alg:censored}
  \SetKwInput{KwInput}{Input}
  \DontPrintSemicolon
  \KwInput{abstention threshold $\rho\in(0,\nicefrac12]$ and learning rate $\eta>0$.}
  \For{$t\in I_r$}{
    Receive $x_t$ from the adversary, and compute $a_t=\sum_{b\in\mathcal B_r}p_t(E_b)\1\{E_b(x_t)=\perp\}$.\;
    \If{$a_t>\rho$}{
      Output an abstention $\hat y_t=\perp$, receive no label, and set $p_{t+1}=p_t$.\;
    }
    \Else{
      Sample a random expert $E_b\sim p_t(\cdot\mid E_b(x_t)\neq\perp)$ and predict $\hat y_t=E_b(x_t)\in\{0,1\}$.\;
      Upon receiving the feedback label $y_t$, update the mixture $p_t$:
      \[p_{t+1}(E_b) = p_t(E_b)\exp\left( -\eta \1\{E_b(x_t) = 1 - y_t\} \right) / Z_t,\]
      where $Z_t$ is a normalization constant.\;
    }
  }
\end{algorithm2e}

\paragraph{Differences from the uncensored setting.} Let $J_r:=\set{t\in I_r: a_t\leq\rho}$ denote the (random) set of rounds in block $r$ where \prettyref{alg:censored} does not abstain, and let $\eps_r:=O(d_\Bc\log T)/m_r$ denote the bound, established in the proof of \prettyref{lem:abstentions}, on the probability that any expert in block $r$ abstains on an i.i.d.~example. The following table summarizes the differences between the two algorithms and their analyses.
\begin{center}
  \small
  \begin{tabular}{lll}
    \toprule
    & \prettyref{alg:agnostic} (\prettyref{thm:main-result}) & \prettyref{alg:censored} (\prettyref{thm:censored}) \\
    \midrule
    Abstains when & the sampled expert abstains & $a_t>\rho$ \\
    Labels observed & in every round & only in the rounds in $J_r$ \\
    Mixture updated on & every round of block $r$ & only the rounds in $J_r$ \\
    Benchmark & $\OPT$ & $\OPTc$ \\
    Misclassification & \prettyref{lem:mis-generic} & \prettyref{lem:cen-mis}: extra factor $1/(1-\rho)$ \\
    Abstention & \prettyref{lem:abstentions}: $\E[a_t]\leq\eps_r$ & \prettyref{lem:cen-abs}: $\Prob(a_t>\rho)\leq\eps_r/\rho$ \\
    \bottomrule
  \end{tabular}
\end{center}
The extra factor $1/(1-\rho)$ in the misclassification error bound arises because \prettyref{alg:censored} must predict even when some experts abstain, and the extra factor $1/\rho$ in the abstention error bound arises because it abstains based on the threshold $\rho$ rather than on the prediction of the sampled expert.

\begin{lemma}[Misclassification Error under Censored Feedback]\label{lem:cen-mis}
  For any distribution $D$ over $\X$, any (possibly adaptive) adversary, and any $\rho\in(0,\nicefrac12]$, \prettyref{alg:censored} with $\eta=\sqrt{8\log|\mathcal B_r|/\tau}$ in each block $r$ satisfies
  \[
    \E\brk{\Mis}\leq\frac{1}{1-\rho}\prn{\E\brk{\OPTc}+\sqrt{(T^2/2\tau)\cdot\log N}}.
  \]
  Moreover, in the realizable case where $\OPTc=0$, \prettyref{alg:censored} with $\eta\geq\log(2)$ satisfies $\E\brk{\Mis}\leq\frac{1}{1-\rho}\cdot 2(T/\tau)\cdot\log N$.
\end{lemma}

\begin{proof}
  We follow the proof of \prettyref{lem:mis-generic}, and use its notation $\mathsf H_{t-1}$, $\mathsf H^{+}_{t-1}$, and $\ell_t(E_b)=\1\set{E_b(x_t)=1-y_t}$. Since the mixture $p_t$ is determined by $\mathsf H_{t-1}$, the quantity $a_t$ and the decision whether to abstain are determined by $\mathsf H^{+}_{t-1}$, and $\hat y_t\neq\perp$ if and only if $t\in J_r$, for every block $r\geq 2$ and $t\in I_r$.

  The first difference from the proof of \prettyref{lem:mis-generic} is that, when it does not abstain, \prettyref{alg:censored} samples its expert from $p_t(\cdot\mid E_b(x_t)\neq\perp)$. Since abstaining experts incur no loss, for every $t\in J_r$,
  \[
    \Prob\prn{\hat y_t=1-y_t\mid\mathsf H^{+}_{t-1}}
    =\frac{1}{1-a_t}\sum_{b\in\mathcal B_r}p_t(E_b)\,\ell_t(E_b)
    \leq\frac{1}{1-\rho}\,\E_{E_b\sim p_t}\,\ell_t(E_b).
  \]
  Since \prettyref{alg:censored} abstains on every timestep of block $1$, by the law of total expectation,
  \[
    \E\brk{\Mis}\leq\frac{1}{1-\rho}\,\E\left[\sum_{r=2}^{R}\sum_{t\in J_r}\E_{E_b\sim p_t}\,\ell_t(E_b)\right],
  \]
  and we bound the quantity inside the expectation for every realization of the interaction.

  The second difference is that \prettyref{alg:censored} only updates its mixture on the rounds in $J_r$, and so the mixtures $(p_t)_{t\in J_r}$ are the iterates of Multiplicative Weights on the loss functions $(\ell_t)_{t\in J_r}$. Since $|J_r|\leq\tau$, \prettyref{lem:multiplicativeweights} with $\eta=\sqrt{8\log|\mathcal B_r|/\tau}$ bounds the regret on $J_r$ by $\frac{\log|\mathcal B_r|}{\eta}+\frac{\eta|J_r|}{8}\leq\sqrt{(\tau/2)\log|\mathcal B_r|}$, even though $J_r$ is selected adaptively. Let $h_\star$ be a minimizer in the definition of $\OPTc$ for the realized interaction. As in the proof of \prettyref{lem:mis-generic}, applying \prettyref{lem:dominate} to a bracket of $h_\star$ in $\mathcal B_r$, and using $|\mathcal B_r|\leq N$, gives
  \[
    \sum_{t\in J_r}\E_{E_b\sim p_t}\,\ell_t(E_b)\leq\sum_{t\in J_r}\1\bigl[h_\star(x_t)\neq y_t\bigr]+\sqrt{\frac{\tau}{2}\log N}.
  \]
  Since the rounds where \prettyref{alg:censored} does not abstain are exactly $\bigcup_{r\geq 2}J_r$, summing over $r=2,\dots,R$ bounds the quantity inside the expectation by $\OPTc+\sqrt{(T^2/2\tau)\log N}$, which proves the first claim. For the second claim, $\OPTc=0$ implies that the bracket of $h_\star$ incurs no loss on $J_r$, and we invoke the realizable guarantee of \prettyref{lem:multiplicativeweights} instead.
\end{proof}

\begin{lemma}[Abstention Error under Censored Feedback]\label{lem:cen-abs}
  For any $\rho\in(0,1)$, \prettyref{alg:censored} satisfies
  \[
    \E\brk{\Abs}\leq 2\tau+O\prn{\frac{d_\Bc\log T}{\rho}}\log\prn{\frac{T}{\tau}}.
  \]
\end{lemma}

\begin{proof}
  The proof is identical to that of \prettyref{lem:abstentions}, except for the bound on the probability of abstaining on an i.i.d.~example in a block $r\geq r_\tau$: \prettyref{alg:agnostic} abstains with probability $a_t$, whereas \prettyref{alg:censored} abstains exactly when $a_t>\rho$. Fix a block $r\geq r_\tau$ and a timestep $t\in I_r$. As in the proof of \prettyref{lem:abstentions}, on the event $A_{m_r}$, conditioned on $\mathsf H_{t-1}$ and on $c_t=0$, the example $x_t$ is a fresh draw from $D$ on which every expert $E_b$ with $b\in\mathcal B_r$ abstains with probability at most $\eps_r$, and therefore $\E[a_t\mid\mathsf H_{t-1},c_t=0]\leq\eps_r$. By Markov's inequality,
  \[
    \Prob\prn{\hat y_t=\perp\mid\mathsf H_{t-1},c_t=0}=\Prob\prn{a_t>\rho\mid\mathsf H_{t-1},c_t=0}\leq\frac{\eps_r}{\rho}.
  \]
  Following the rest of the proof of \prettyref{lem:abstentions} with this additional factor of $1/\rho$ proves the claim.
\end{proof}

\begin{proof}[Proof of {\prettyref{thm:censored}}]
  Let $\tau=\ceil{T^{1-\alpha}}$. In the agnostic setting, we run \prettyref{alg:censored} with $\eta=\sqrt{8\log|\mathcal B_r|/\tau}$ in each block $r$, and $\rho=\min\set{\nicefrac12,\sqrt{\log N/T^{1-\alpha}}}$. Then $\rho T\leq\sqrt{T^{1+\alpha}\log N}$: this holds with equality if $\rho<\nicefrac12$, and otherwise $\log N\geq T^{1-\alpha}/4$, and so $\rho T=T/2\leq\sqrt{T^{1+\alpha}\log N}$. Since $\OPTc\leq T$ and $\frac{1}{1-\rho}\leq 2$, \prettyref{lem:cen-mis} gives
  \begin{align*}
    \E\brk{\Mis-\OPTc}
    &\leq\frac{\rho}{1-\rho}\,\E\brk{\OPTc}+\frac{1}{1-\rho}\sqrt{\frac{T^2}{2\tau}\log N}\\
    &\leq 2\rho T+\sqrt{2T^{1+\alpha}\log N}
    \leq 4\sqrt{T^{1+\alpha}\log N}.
  \end{align*}
  Moreover, $\frac{1}{\rho}\leq 2+\sqrt{T^{1-\alpha}/\log N}\leq 2+\sqrt{T^{1-\alpha}}$, since $\log N\geq 1$. Therefore, \prettyref{lem:cen-abs}, $\log(T/\tau)\leq\log T$, and the AM-GM inequality give
  \[
    \E\brk{\Abs}
    \leq 2\tau+O\prn{d_\Bc\log^2 T}\cdot\prn{2+\sqrt{T^{1-\alpha}}}
    \leq 3T^{1-\alpha}+O\prn{d_\Bc^2\log^4 T}.
  \]
  In the realizable setting, we run \prettyref{alg:censored} with $\eta=\log 2$ and $\rho=\nicefrac12$. Then, \prettyref{lem:cen-mis} gives $\E\brk{\Mis}\leq 4(T/\tau)\log N\leq 4T^{\alpha}\log N$, and \prettyref{lem:cen-abs} gives $\E\brk{\Abs}\leq 2T^{1-\alpha}+O\prn{d_\Bc\log^2 T}$.
\end{proof}

\subsection{Censored Guarantees for Some Natural Function Classes}

\begin{corollary}[Halfspaces]\label{corr:cen-halfspaces}
  Let $\Hc$ be halfspaces in $\mathbb{R}^d$. Let $T\geq 1$ be a time horizon. Then, for any parameter $\alpha \in [0,1]$, \prettyref{alg:censored} achieves the following guarantee against {\em any} adversary:
  \[
    \E[\Mis - \OPTc] \lesssim \sqrt{T^{1+\alpha}\cdot d\log(2dT)} \text{ and }\E[\Abs] \lesssim T^{1-\alpha} + d^{4}\log^{2}(d)\log^{4}(T).
  \]
  Moreover, in the realizable setting where $\OPTc=0$, \prettyref{alg:censored} satisfies
  \[
    \E[\Mis] \lesssim T^{\alpha} \cdot d\log(2dT) \text{ and } \E[\Abs] \leq 2T^{1-\alpha} + O(d^{2}\log(d)\log^{2}(T)).
  \]
\end{corollary}

\begin{corollary}[Compositions of Halfspaces]
  \label{corr:cen-comp-halfspaces}
  Let $\Gc=\{f(h_1, \dots, h_k): f\in \Fc, h_1,\dots, h_k \in \Hc \}$, where $\Hc$ is the class of halfspaces in $\mathbb{R}^d$ and $\Fc$ is an {\em arbitrary} function class with VC dimension $d_\Fc$. Let $T\geq 1$ be a time horizon. Then, for any parameter $\alpha \in [0,1]$, \prettyref{alg:censored} achieves the following guarantee against {\em any} adversary:
  \begin{align*}
    \E[\Mis - \OPTc] &\lesssim \sqrt{T^{1+\alpha}\cdot k(d_\Fc+d\log(2dkT))}, \text{ and}\\
    \E[\Abs] &\lesssim T^{1-\alpha} + k^{2}d^{4}\log^{2}(d)\log^{4}(T).
  \end{align*}
  Moreover, in the realizable setting where $\OPTc=0$, \prettyref{alg:censored} satisfies
  \[
    \E[\Mis] \lesssim T^{\alpha} \cdot k(d_\Fc+d\log(2dkT)) \text{ and }\E[\Abs] \leq 2T^{1-\alpha} + O(kd^{2}\log(d)\log^{2}(T)).
  \]
\end{corollary}

\section{Discussion}
\label{sec:conclusion}

In this work, we present new algorithms and learning guarantees for online classification with abstention under adversarial injections \cite{DBLP:conf/nips/GoelHMS23}, in the agnostic and realizable settings, against oblivious and adaptive adversaries, for function classes that admit bounded bracketing numbers witnessed by brackets of bounded VC dimension. We discuss below a few directions for future research.

\paragraph{VC Dimension vs.~Bracketing Number.} Our approach fundamentally relies on the concept of brackets and is limited to function classes with bounded bracketing number. A natural technical question that arises in our work is: if a function class $\Hc$ with VC dimension $d$ has finite $\eps$-bracketing number satisfying $\Nbr{\Hc}{D}{\eps}\leq (1/\eps)^{O(d)}$, then does there exist a class $\Bc$ with VC dimension ${\rm poly}(d)$ that brackets $\Hc$ (as in \prettyref{def:bracket-class})? In other words, whenever small bracketing number is possible, is it possible to achieve with brackets with low complexity? This question does not appear to be studied in the literature, and it would strengthen recent results \citep*{DBLP:conf/innovations/DuttaGM22} bounding the bracketing number of several natural function classes if resolved positively.

\paragraph{Guarantees for VC classes against adaptive adversaries.} While our results establish new guarantees against adaptive adversaries in both the realizable and agnostic settings for function classes with bounded bracketing numbers (including halfspaces and feedforward nets), as discussed earlier, bracketing numbers do not behave ``nicely'' for every VC class. Therefore, our approach does not fully resolve the open question of whether one can achieve sublinear ${\rm poly}(d)T^{1 - \Omega(1)}$ misclassification and abstention error rates against adaptive adversaries for every function class of VC dimension $d$ \cite{DBLP:conf/nips/GoelHMS23, YB26}.

\paragraph{Computational aspects.} The runtime of our bracketing-based algorithm is linear in the bracketing number which for halfspaces in $\mathbb{R}^d$ scales as $n^{O(d)}$ where $n$ is the size of the (contaminated) sequence. We remark that this is a huge improvement over previous approaches which require calculating involved shattering statistics \cite{DBLP:conf/nips/GoelHMS23} and running exponentially many algorithmic instantiations \cite{YB26}. It would be interesting to explore more efficient algorithms for halfspaces. More broadly, it would also be interesting to think about oracle-efficient algorithms with access to reasonable oracles (e.g., ERM).

\section*{Acknowledgments}

The author thanks the anonymous reviewers for their feedback, which helped improve the presentation of this paper.

\section*{Statement on AI Usage}

The central conceptual ideas of this paper are: (1) the agnostic formulation of online learning with abstention under adversarial injections, and (2) the use of the machinery of brackets and bracketing numbers to tackle this problem. Both were conceived and developed by the author without the use of AI tools.

In later stages, the author used ChatGPT 5.4 Pro to explore how brackets for halfspaces can be constructed from i.i.d.~samples, and asked whether such a construction remains possible when the samples are adversarially contaminated. It identified that uniform convergence can be used to guarantee this. The author then generalized this observation from halfspaces to generic function classes, which led to \prettyref{def:bracket-class} and \prettyref{lem:contam}. ChatGPT 5.5 Pro was used to help with the proofs for censored feedback (\prettyref{sec:censored}), which followed directly from \prettyref{thm:main-result} and its proof (\prettyref{sec:main-proof}). Finally, much later, while preparing the final version of this paper, Claude Opus 5.5 was used to polish some parts of the writing.

\bibliographystyle{surnameinitials}
\bibliography{refs}

\newpage

\appendix

\section{Warmup: Known Marginal Distribution}
\label{app:known-d-baseline}

For completeness, we formalize here the simple known-distribution warmup discussed in the technical overview.

\begin{algorithm2e}[H]
  \caption{Known-distribution learning via fixed bracket experts}
  \label{alg:known-dist}
  \SetKwInput{KwInput}{Input}
  \DontPrintSemicolon
  \KwInput{horizon $T$, hypothesis class $\Hc$, distribution $D$ over $\X$, and scale $\eps\in(0,1)$.}
  Let $\mathcal B_{D,\eps}$ be an $\eps$-bracketing family for $\Hc$ with respect to $D$ such that $|\mathcal B_{D,\eps}| = \Nbr{\Hc}{D}{\eps}$.\;
  For each bracket $b=(b^-,b^+)\in \mathcal B_{D,\eps}$, define an abstaining expert
  \[
    E_b(x):= b^-(x) \text{ if } b^-(x)=b^+(x), \text{ otherwise }\perp.
  \]\;
  Initialize $p_1$ to be a uniform distribution over $\{E_b:b\in \mathcal B_{D,\eps}\}$.\;
  \For{$t=1,\dots,T$}{
    Receive $x_t$ from the adversary.\;
    Sample a random expert $E_b\sim p_t$ and predict $\hat y_t = E_b(x_t)\in\{0,1,\perp\}$.\;
    Upon receiving the label $y_t$, update the mixture:
    \[
      p_{t+1}(E_b) = p_t(E_b)\exp\left(-\eta\1\{E_b(x_t)=1-y_t\}\right)/Z_t,
    \]
    where $Z_t$ is a normalization constant.\;
  }
\end{algorithm2e}

\begin{theorem}
  \label{thm:known-dist}
  Let $\Hc$ be a hypothesis class, let $D$ be a distribution on $\X$, and let $\eps\in(0,1)$. Then, against any (possibly adaptive) adversary, \hyperref[alg:known-dist]{Algorithm~\ref*{alg:known-dist}} with parameter $\eta=\sqrt{8\log \Nbr{\Hc}{D}{\eps}/T}$ guarantees
  \[
    \E[\Mis - \OPT]
    \leq
    \sqrt{(T/2)\log \Nbr{\Hc}{D}{\eps}}
    \qquad\text{and}\qquad
    \E[\Abs] \leq \eps T.
  \]
\end{theorem}

\begin{proof}
  Write $\mathcal B:=\mathcal B_{D,\eps}$ and $N:=|\mathcal B|=\Nbr{\Hc}{D}{\eps}$. We use the notation $\mathsf H_{t-1}$, $\mathsf H^{+}_{t-1}$, and $\ell_t(E_b)=\1\set{E_b(x_t)=1-y_t}$ from the proof of \prettyref{lem:mis-generic}, with \prettyref{alg:known-dist} in place of \prettyref{alg:agnostic}. Since $\mathcal B$ is fixed in advance, the mixture $p_t$ is determined by $\mathsf H_{t-1}$, and at time $t$ \prettyref{alg:known-dist} samples $E_b\sim p_t$ using fresh randomness that is independent of $\mathsf H^{+}_{t-1}$.

  \paragraph{Misclassification error.} As in the proof of \prettyref{lem:mis-generic}, the conditional expected misclassification error at time $t$ given $\mathsf H^{+}_{t-1}$ is $\E_{E_b\sim p_t}\ell_t(E_b)$, and therefore, by the law of total expectation,
  \[
    \E[\Mis] = \E\brk{\sum_{t=1}^{T}\E_{E_b\sim p_t}\ell_t(E_b)}.
  \]
  We bound the quantity inside the expectation for every realization of the interaction. Fix such a realization, and let $h_\star\in\arg\min_{h\in\Hc}\sum_{t=1}^{T}\1\set{h(x_t)\neq y_t}$ for the realized sequence. Since $\mathcal B$ is an $\eps$-bracketing of $\Hc$, it contains a bracket $b_\star$ for $h_\star$, and \prettyref{lem:dominate} gives $\ell_t(E_{b_\star})\leq\1\set{h_\star(x_t)\neq y_t}$ for every $t$. Combining this with the regret guarantee of Multiplicative Weights (\prettyref{lem:multiplicativeweights}), which holds for every sequence of loss functions, gives
  \[
    \sum_{t=1}^{T}\E_{E_b\sim p_t}\ell_t(E_b)
    \leq
    \sum_{t=1}^{T}\ell_t(E_{b_\star}) + \sqrt{(T/2)\log N}
    \leq
    \OPT + \sqrt{(T/2)\log N}.
  \]
  Taking expectations proves the bound on $\E[\Mis-\OPT]$.

  \paragraph{Abstention error.} Since $\mathcal B$ has minimum size, every bracket $b=(b^-,b^+)\in\mathcal B$ has disagreement mass $\Prob_{X\sim D}(b^-(X)\neq b^+(X))\leq\eps$ (otherwise it could be removed from $\mathcal B$). Moreover, $E_b(x)=\perp$ exactly when $b^-(x)\neq b^+(x)$. Fix a round $t$. Conditioned on $\mathsf H_{t-1}$ and on $c_t=0$, the example $x_t$ is a fresh draw from $D$ that is independent of the expert $E_b\sim p_t$ sampled by \prettyref{alg:known-dist}, and therefore
  \[
    \Prob\prn{\hat y_t=\perp \mid \mathsf H_{t-1}, c_t=0}
    =
    \sum_{b\in\mathcal B}p_t(E_b)\,\Prob_{X\sim D}\prn{b^-(X)\neq b^+(X)}
    \leq
    \eps.
  \]
  Hence, $\Prob(c_t=0\text{ and }\hat y_t=\perp\mid\mathsf H_{t-1})\leq\eps\,\Prob(c_t=0\mid\mathsf H_{t-1})$. Summing over $t$ and taking expectations gives
  \[
    \E[\Abs]
    =
    \sum_{t=1}^{T}\Prob\prn{c_t=0\text{ and }\hat y_t=\perp}
    \leq
    \eps\,\E\brk{\sum_{t=1}^{T}\1\set{c_t=0}}
    \leq
    \eps T.
  \qedhere
  \]
\end{proof}

\section{Bracketing Number of Compositions}
\label{app:bracket}

\compositionlemma*

\begin{remark}\label{rem:compo-samples}
  The construction used below in the proof of \prettyref{lem:composition-bracketing} satisfies the following additional property. For any integer $m\geq 1$, if for each $i\in [k]$ a family of brackets for $\Hc_i$ can be constructed from $m$ i.i.d.~samples from $D$ (that may be adversarially contaminated as in \prettyref{lem:contam}) with scale $\eps_i(m,\delta_i)$ with probability at least $1-\delta_i$, then a family of brackets for the composition $\Gc$ can be constructed from $m$ i.i.d.~samples from $D$ (that may be adversarially contaminated as in \prettyref{lem:contam}) with scale $\sum_{i}\eps_i(m,\delta_i)$ with probability at least $1-\sum_i \delta_i$.

  For example, when each $\Hc_i$ is halfspaces in $\mathbb{R}^d$, the bracket for $\Gc$ will have scale proportional to $k \cdot\frac{O(d^2\log(d)\log(m) +\log(1/\delta))}{m}$ implying a final abstention error of \prettyref{alg:agnostic} with an additive term proportional to $O(kd^2\log(d)\log^2(T))$ which can be smaller than the VC dimension of $\Gc$ (or its underlying bracketing class).
\end{remark}

\begin{proof}[Proof of {\prettyref{lem:composition-bracketing}}]
  If $\Nbr{\Hc_i}{D}{\eps_i}=\infty$ for some $i\in[k]$, there is nothing to prove, so assume that all of these bracketing numbers are finite. For each $i\in[k]$, let $\mathcal B_i$ be an $\eps_i$-bracketing of $\Hc_i$ with respect to $D$ of size $|\mathcal B_i|=\Nbr{\Hc_i}{D}{\eps_i}$.

  \paragraph{Construction.} For a tuple of brackets $b=(b_1,\dots,b_k)\in\mathcal B_1\times\cdots\times\mathcal B_k$ with $b_i=(b_i^-,b_i^+)$ and a point $x\in\X$, let
  \[
    Q_b(x) := \set{z\in\{0,1\}^k :\ b_i^-(x)\le z_i\le b_i^+(x) \text{ for all } i\in[k]}
  \]
  be the set of label vectors that are consistent with the brackets $b_1,\dots,b_k$ at $x$. Since $b_i^-(x)\le b_i^+(x)$ for every $i\in[k]$, the set $Q_b(x)$ is nonempty. For each $f\in\Fc$, define
  \[
    g^-_{b,f}(x) := \min_{z\in Q_b(x)} f(z)
    \qquad\text{and}\qquad
    g^+_{b,f}(x) := \max_{z\in Q_b(x)} f(z).
  \]
  Clearly $g^-_{b,f}(x)\le g^+_{b,f}(x)$ for every $x\in\X$, so $(g^-_{b,f},g^+_{b,f})$ is a bracket. We will show that
  \[
    \mathcal B_\Gc := \set{(g^-_{b,f},g^+_{b,f}) :\ b\in\mathcal B_1\times\cdots\times\mathcal B_k,\ f\in\Fc}
  \]
  is an $(\eps_1+\cdots+\eps_k)$-bracketing of $\Gc$ with respect to $D$.

  \paragraph{Bracketing.} Fix $g=f(h_1,\dots,h_k)\in\Gc$ with $f\in\Fc$ and $h_i\in\Hc_i$. For each $i\in[k]$, let $b_i=(b_i^-,b_i^+)\in\mathcal B_i$ be a bracket of $h_i$ with $\Prob_{X\sim D}(b_i^-(X)\neq b_i^+(X))\le\eps_i$, and let $b=(b_1,\dots,b_k)$. For every $x\in\X$, we have $b_i^-(x)\le h_i(x)\le b_i^+(x)$ for all $i\in[k]$, that is, $(h_1(x),\dots,h_k(x))\in Q_b(x)$, and therefore
  \[
    g^-_{b,f}(x)\le f\bigl(h_1(x),\dots,h_k(x)\bigr) = g(x) \le g^+_{b,f}(x).
  \]
  Hence, $(g^-_{b,f},g^+_{b,f})$ brackets $g$. Moreover, if $b_i^-(x)=b_i^+(x)$ for all $i\in[k]$, then $Q_b(x)$ is a singleton, and so $g^-_{b,f}(x)=g^+_{b,f}(x)$. Thus,
  \[
    \set{x\in\X :\ g^-_{b,f}(x)\neq g^+_{b,f}(x)} \subseteq \bigcup_{i=1}^{k}\set{x\in\X :\ b_i^-(x)\neq b_i^+(x)},
  \]
  and by a union bound,
  \[
    \Prob_{X\sim D}\prn{g^-_{b,f}(X)\neq g^+_{b,f}(X)} \le \sum_{i=1}^{k}\Prob_{X\sim D}\prn{b_i^-(X)\neq b_i^+(X)} \le \sum_{i=1}^{k}\eps_i.
  \]

  \paragraph{Counting.} Since $\Fc$ is a class of functions over the domain $\{0,1\}^k$ of size $2^k$, the Sauer--Shelah--Perles lemma gives $|\Fc|\le\sum_{j=0}^{d_\Fc}\binom{2^k}{j}\le\prn{e2^k/d_\Fc}^{d_\Fc}\le e\,2^{kd_\Fc}$ when $d_\Fc\geq 1$, where the last inequality uses $(e/d)^d\le e$ for every $d\geq 1$ (and $|\Fc|\le 1$ when $d_\Fc=0$). Therefore,
  \[
    \Nbr{\Gc}{D}{\eps_1+\cdots+\eps_k}
    \le |\mathcal B_\Gc|
    \le |\Fc|\prod_{i=1}^{k}|\mathcal B_i|
    \le e\,2^{kd_\Fc}\prod_{i=1}^{k}\Nbr{\Hc_i}{D}{\eps_i}.
  \qedhere
  \]
\end{proof}

\end{document}